\documentclass{article}
\usepackage{PRIMEarxiv}

\usepackage{cite}
\usepackage{amsmath,amssymb,amsfonts}
\usepackage{algorithmic}
\usepackage{graphicx}
\usepackage{textcomp}
\def\BibTeX{{\rm B\kern-.05em{\sc i\kern-.025em b}\kern-.08em
    T\kern-.1667em\lower.7ex\hbox{E}\kern-.125emX}}

\usepackage{xcolor, soul, tikz, svg, subcaption, enumitem, adjustbox, float}
\usepackage{array, caption, makecell, multirow, tabularx, tcolorbox, verbatim}
\usepackage[utf8]{inputenc} % allow utf-8 input
\usepackage[T1]{fontenc}    % use 8-bit T1 fonts
\usepackage{hyperref}       % hyperlinks
\usepackage{url}            % simple URL typesetting
\usepackage{booktabs}       % professional-quality tables
\usepackage{amsfonts}       % blackboard math symbols
\usepackage{nicefrac}       % compact symbols for 1/2, etc.
\usepackage{microtype}      % microtypography
\usepackage{lipsum}
\usepackage{fancyhdr}       % header
\usepackage{graphicx}       % graphics
\graphicspath{{media/}}     % organize your images and other figures under media/ folder

\definecolor{lime}{HTML}{A6CE39}
\DeclareRobustCommand{\orcidicon}{
	\begin{tikzpicture}
	\draw[lime, fill=lime] (0,0) 
	circle [radius=0.16] 
	node[white] {{\fontfamily{qag}\selectfont \tiny ID}};
	\draw[white, fill=white] (-0.0625,0.095) 
	circle [radius=0.007];
	\end{tikzpicture}
	\hspace{-2mm}
}
\foreach \x in {A, ..., Z}{\expandafter\xdef\csname orcid\x\endcsname{\noexpand\href{https://orcid.org/\csname orcidauthor\x\endcsname}
			{\noexpand\orcidicon}}
}

\title{Modeling and Optimizing User Preferences in AI Copilots: A Comprehensive Survey and Taxonomy
}

\author{
    Saleh Afzoon $^{1}$ \orcidA{}, 
    Ali Shahsavandi $^{1}$ \orcidB{},
    Phuong Thao Huynh $^{1}$ \orcidC{},
    Melika Zare $^{1}$ \orcidD{},\\
    \textbf{Zahra Jahanandish} $^{2}$ \orcidE{},
    \textbf{Amin Beheshti} $^{1}$\orcidF{},
    \textbf{Usman Naseem} $^{1}$ \orcidG{}\\\\
    $^{1}$School of Computing, Macquarie University, Sydney, Australia \\
    \small \texttt{\{saleh.afzoon, ali.shahsavandi, melika.zare\}@hdr.mq.edu.au},\\ \small \texttt{\{jasmine.huynh, amin.beheshti, usman.naseem\}@mq.edu.au},\\
    $^{2}$Department of Computer Engineering and Information Technology, Shiraz University of Technology, Shiraz, Iran \\
    \small \texttt{zjahanandish@gmail.com},
}

\begin{document}
\maketitle

% \begin{document}

% \title{Modeling and Optimizing User Preferences in AI Copilots: A Comprehensive Survey and Taxonomy}

% \author{Saleh Afzoon}
% \email{saleh.afzoon@hdr.mq.edu.au}
% \authornote{Corresponding author}
% \orcid{0000-0002-4253-9069}
% \affiliation{%
%   \institution{Macquarie University}
%   \city{Sydney}
%   \state{NSW}
%   \country{Australia}
% }

% \author{Ali Shahsavandi}
% \email{ali.shahsavandi@hdr.mq.edu.au}
% \orcid{0009-0004-6190-8976}
% \affiliation{%
%   \institution{Macquarie University}
%   \city{Sydney}
%   \state{NSW}
%   \country{Australia}
% }
% \author{Phuong Thao Huynh}
% \orcid{0009-0003-1946-6863}
% \email{jasmine.huynh@mq.edu.au}
% \affiliation{%
%   \institution{Macquarie University}
%   \city{Sydney}
%   \state{NSW}
%   \country{Australia}
% }
% \author{Melika Zare}
% \email{melika.zare@hdr.mq.edu.au}
% \orcid{0009-0008-1588-401X}
% \affiliation{%
%   \institution{Macquarie University}
%   \city{Sydney}
%   \state{NSW}
%   \country{Australia}
% }
% \author{Zahra Jahanandish}
% \orcid{0009-0004-0406-8886}
% \email{zjahanandish@gmail.com}
% \affiliation{%
%   \institution{Shiraz University of Technology}
%   \city{Shiraz}
%   \state{Fars}
%   \country{Iran}
% }

% \author{Amin Beheshti}
% \orcid{0000-0002-5988-5494}
% \email{amin.beheshti@mq.edu.au}
% \affiliation{%
%   \institution{Macquarie University}
%   \city{Sydney}
%   \state{NSW}
%   \country{Australia}
% }

% \author{Usman Naseem}
% \orcid{0000-0003-0191-7171}
% \email{usman.naseem@mq.edu.au}
% \affiliation{%
%   \institution{Macquarie University}
%   \city{Sydney}
%   \state{NSW}
%   \country{Australia}
% }

\begin{abstract}

AI copilots represent a new generation of AI-powered systems designed to assist users, particularly knowledge workers and developers, in complex, context-rich tasks. As these systems become more embedded in daily workflows, personalization has emerged as a critical factor for improving usability, effectiveness, and user satisfaction. Central to this personalization is preference optimization: the system’s ability to detect, interpret, and align with individual user preferences. While prior work in intelligent assistants and optimization algorithms is extensive, their intersection within AI copilots remains underexplored.
This survey addresses that gap by examining how user preferences are operationalized in AI copilots. We investigate how preference signals are sourced, modeled across different interaction stages, and refined through feedback loops. Building on a comprehensive literature review, we define the concept of an AI copilot and introduce a taxonomy of preference optimization techniques across pre-, mid-, and post-interaction phases. Each technique is evaluated in terms of advantages, limitations, and design implications.
By consolidating fragmented efforts across AI personalization, human-AI interaction, and language model adaptation, this work offers both a unified conceptual foundation and a practical design perspective for building user-aligned, persona-aware AI copilots that support end-to-end adaptability and deployment.

\end{abstract}


\section{Introduction}

%%% 1- Context, 2- Its importance
AI-powered, context-aware systems, commonly referred to as AI copilots, are rapidly emerging as collaborative assistants for developers, analysts, and knowledge workers engaged in complex, high-context tasks. These systems are integrated into user workflows and designed to understand evolving goals, adapt to individual preferences, and provide real-time, personalized support. However, achieving meaningful personalization in such systems presents unique challenges. User intent is often ambiguous or changes over time, tasks may span multiple domains, and responses must be dynamically adapted to nuanced contexts. These systems must be capable of aligning with dynamic user preferences, as personalization is a key success factor for digital assistants, enabling them to adapt their responses and behavior to individual users’ routines and context \cite{maedche2019ai}.
At the core of this personalization lies preference optimization, the ability of a system to detect, interpret, and align with user preferences \cite{xiao2024comprehensive}. While advances in preference modeling have gained traction in machine learning and recommender systems, their integration into real-time, interactive systems like AI copilots remains limited and fragmented. Designing AI copilots that can continuously adapt to evolving preferences in high-context environments is still a major open challenge.

This survey is driven by the need to systematically examine how preference optimization is realized in AI copilots. Our aim is to consolidate disparate research across assistant technologies, personalization, and alignment learning, and to provide a structured account of how preferences are acquired, modeled, and refined in these systems. In doing so, we offer a comprehensive view of the emerging design space for adaptive, user-centered AI copilots.

Research related to this survey broadly falls into two categories: (i) AI systems and intelligent assistants, and (ii) preference optimization techniques. While each has been explored independently in the literature, their integration remains underexamined.
The evolution of Virtual Assistants (VAs), such as Google Assistant, Siri, Cortana, and Alexa, has been well documented, with studies focusing on their core functionalities, user experience, and limitations in areas such as voice recognition, contextual awareness, and multi-turn dialogue management \cite{tulshan2019survey}. In enterprise contexts, Digital Assistants (DAs) have been examined through the lens of business integration, workflow support, and human-computer collaboration, highlighting both their organizational potential and associated risks \cite{maedche2019ai}. Concerns around privacy and security in these systems have also been widely addressed, particularly in relation to data handling and unauthorized access \cite{bolton2021security}. Although the design and evaluation of AI copilots in retail has been recently analyzed through Microsoft’s case studies, the work is limited in scope and does not offer a comprehensive perspective on the broader cognitive assistant landscape \cite{furmakiewicz2024design}.
On the other hand, preference optimization, especially in the context of aligning AI behavior with user intent, has seen significant theoretical and practical development. A recent comprehensive survey on Direct Preference Optimization (DPO) presents an in-depth analysis of its theoretical foundations, training methodologies, benchmark datasets, and diverse application scenarios \cite{xiao2024comprehensive}. However, such works primarily examine preference modeling in isolation and do not address its integration within interactive systems like AI copilots or digital assistants.

%%% 4- what previous studies lack
While existing studies contribute significantly to our understanding of intelligent assistant technologies and preference optimization methods, their integration in the context of AI copilots remains limited and underexplored. As a result, there is no consolidated view of how preference signals can be effectively incorporated into the behavior of real-time, context-aware systems.
This survey aims to bridge these two streams of research by examining preference optimization techniques within the design of AI copilots, systems that operate collaboratively with users in high-context, evolving environments. Specifically, we address how preferences can be detected, modeled, and refined to enhance personalization and system responsiveness. The main contributions of this work are summarized as follows:

%%% 5- List of specific contributions

\begin{itemize}
\item A unified and literature-grounded definition of an AI copilot is proposed, synthesizing diverse descriptions and terminology from recent studies across domains.

\item A comprehensive analytical review of preference optimization in AI copilot design is provided, covering possible sources of user preferences, techniques for preference-aware response generation, and methods for feedback-driven adaptation.
\end{itemize}

%%% 7- Paper's structure
We organized the content of our survey as follows: Section 2 provides foundational background by examining the historical development and evolving terminology associated with AI copilots. Section 3 introduces a unified, literature-informed definition of AI copilots, integrating insights from diverse fields. Section 4 presents a conceptual AI copilot architecture focused on preference optimization, offering a structured analysis of preference signal sources, detection techniques, persona modeling, personalized response generation, evaluation frameworks, and feedback-driven alignment methods. Finally, Section 5 concludes with a summary of key insights and directions for future research.

%%%%%%%%%%%%%%%%%%%%%%%%%%%%%%%%%%%%%%%%%%%%%%%%%%%%%%%%%%%%%%%%%%%%%%%

\section{Preliminary and Background}
\label{sec:background}

A precise definition of AI copilots requires first establishing how related system types have been characterized in the literature. The terminology surrounding AI-powered assistive systems, reviewed below, provides the necessary conceptual context for situating AI copilots within the broader landscape of intelligent assistants.

\subsection{AI Copilot Related Terms}

\subsubsection{Cognitive Assistant.}

Cognitive computing refers to hardware and software systems that emulate human cognitive processes~\cite{kelly2015computing}. \textit{Metacognition} extends this paradigm by enabling self-monitoring and adaptive reasoning, producing symbolic, closed-loop architectures in which humans and machines exchange information through real-time feedback cycles~\cite{schmorrow2005foundations}.
Building on this foundation, the \textit{Joint Cognitive System} framework conceptualizes human and machine agents as goal-driven collaborators that dynamically share cognitive responsibilities within context-specific environments~\cite{hollnagel2005joint}, shifting the focus from automation toward cooperative cognition and mutual adaptation.

Cognitive Assistant (CA) operationalizes these principles by merging human creativity with computational capabilities to enhance problem-solving, decision-making, and situational awareness~\cite{reverberi2022experimental}. Studies in hybrid intelligence highlight their potential to facilitate innovation and support effective teamwork in domain-specific professional contexts~\cite{dellermann2021future, sharma2023human}. IBM Watson Health illustrates this approach, applying cognitive computing to clinical decision support through probabilistic reasoning over structured medical knowledge.

%%%%
\subsubsection{Digital Assistant}

The term Digital Assistant (DA) is frequently interchanged with Virtual Assistant (VA), yet it represents a broader concept. DAs are characterized as versatile, amorphous compound technologies, with mobility, rapid data analysis, and support for multiple interfaces among their key features \cite{wellsandt2021anatomy}. A defining characteristic is their capacity to combine natural language dialogue with personalized, context-specific assistance \cite{knote2019classifying}. Within this broader category, \textit{Smart Personal Assistants} (SPAs) form a personalized subset, further distinguished by interaction modality into text-based and voice-based interfaces~\cite{knote2019classifying, cowan2017can}.

%%%% 
\subsubsection{Virtual Assistant}
To address the need for hands-free, conversational interaction in everyday computing environments, a class of voice-based DAs has emerged, commonly referred to as Virtual Assistants (VAs). Representative systems in this category include Apple Siri and Amazon Alexa, which have achieved broad consumer adoption~\cite{maedche2019ai, acosta2022survey}. Defining features include hands-free operation, context understanding, and the ability to handle multi-turn conversational exchanges~\cite{wellsandt2021anatomy}. Beyond recognizing voice commands and communicating over the same channel~\cite{bolton2021security}, voice-activation capability makes these systems well-suited for integration into everyday devices such as smartphones~\cite{rawassizadeh2019manifestation}.  Text-based interaction is also supported as an additional modality alongside the primary voice interface~\cite{guzman2019voices}, positioning VAs within the broader DA spectrum as systems that are voice-first but not voice-only. In some studies, the same systems are alternatively referred to as SPAs, a label that foregrounds their personalization capabilities rather than their voice interface~\cite{knote2019classifying}.

\subsubsection{Chatbot}

Chatbots represent a foundational category of specialized AI programs, traditionally tailored 
for customer service, information retrieval, and guided interactions within bounded task 
domains~\cite{almansor2020survey}. A defining feature is their capacity to simulate 
conversational exchanges using natural language processing (NLP) and machine learning algorithms, 
creating the impression of dialogue with a human interlocutor~\cite{agarwal2020review, 
hussain2019survey}. Chatbots are primarily classified by their application domain, falling into 
either task-oriented or non-task-oriented categories~\cite{almansor2020survey, hussain2019survey}. 
Task-oriented systems apply domain-specific rules within a focused conversational context, while 
non-task-oriented systems handle open-ended dialogue without a predefined 
goal~\cite{adamopoulou2020chatbots}.

Architecturally, these systems range from retrieval-based models that rank candidate responses 
by similarity to generative models that produce responses word-by-word, with Retrieval-Augmented 
Generation (RAG) emerging as a hybrid approach that combines both 
techniques~\cite{agarwal2020review, siriwardhana2023improving}. A key limitation, however, is 
that chatbots operate under a strictly reactive paradigm: they respond only when prompted via 
text, and rely predominantly on the immediate conversation history rather than maintaining 
awareness of a user's broader digital workspace. Commercial deployments, such as standalone 
customer service bots, illustrate their utility in isolated interactions, distinguishing them 
from embedded, proactive workflow assistants.

\subsubsection{Conversational Agent}

The terms Conversational Agent, Dialogue System, and AI Agent are frequently conflated in the 
literature, yet they represent distinct operational paradigms~\cite{agarwal2020review, 
adamopoulou2020chatbots}. Conversational agents act as reactive dialogue partners within 
bounded interaction contexts, sharing much of their architectural foundation with advanced 
chatbots. Their primary function is to sustain coherent, multi-turn exchanges, typically through 
natural language, within a defined task scope. Unlike chatbots, however, conversational agents 
are designed to manage dialogue state across turns, handle clarification sub-dialogues, and 
adapt responses based on accumulated conversational context~\cite{agarwal2020review}.

A key architectural distinction lies in how dialogue is managed. Retrieval-based conversational 
agents select responses from a predefined candidate set, while generative systems produce 
responses conditioned on the full dialogue history. More recently, Large Language Model (LLM)-based conversational 
agents have extended this capability by incorporating tool use and memory modules, blurring the 
boundary between conversational agents and the broader class of AI agents. AI agents, as 
defined by Wooldridge et al.~\cite{wooldridge1995intelligent}, are goal-driven systems capable 
of autonomous multi-step execution across applications, operating largely without continuous 
human prompting.

Both paradigms, however, present limitations in collaborative knowledge work settings. 
Conversational agents are constrained by their reactive nature and bounded task scope, while 
autonomous AI agents operate outside the human-in-the-loop paradigm. This gap between systems 
that only converse and systems that act independently motivates the AI copilot architecture 
discussed in the following section.

The boundaries between these system types are not always clearly drawn in the literature. Overlapping characterizations reflect the evolving and interdisciplinary nature of the field rather than a settled consensus. Section~\ref{sec:definition} develops a precise, criteria-based definition of AI copilot, as the system class examined in this survey.

\section{AI Copilot}
\label{sec:definition}

The system types reviewed in Section~\ref{sec:background} share surface-level similarities in their use of natural language for user assistance, yet they differ substantially in architecture, autonomy, and adaptability. To establish a precise scope for this survey, this section synthesizes a unified definition of AI copilots from recent literature and positions them against related system types through a structured comparison across five axes.

\subsection{Definition}

The term AI copilot has been applied inconsistently across domains in recent literature, 
describing systems as varied as architectures integrating large language models with retrieval 
mechanisms and responsible AI safeguards~\cite{furmakiewicz2024design}, domain-adapted systems 
supporting clinical decision-making~\cite{lu2024multimodal}, and collaborative partners 
augmenting human capabilities under human supervision~\cite{daugherty2018human+, sellen2024rise}. 
While these characterizations are not mutually exclusive, none individually captures the 
operational properties that distinguish AI copilots from related system types.

Synthesizing these perspectives, an \textbf{AI copilot} is defined as a preference-aware 
collaborative system that assists knowledge workers interactively in real time, operating within 
the user's primary workflow environment under continuous human oversight. Preference awareness, 
in this context, is task-driven and contextualized: rather than adapting to general 
conversational style or tone, an AI copilot models how the user wants a specific task executed 
within a particular artifact or workspace, and refines this understanding as the task evolves.

Four operational characteristics jointly distinguish AI copilots from related system types: inline 
delivery of assistance within the user's primary workflow environment; contextual grounding in 
the surrounding artifact rather than the conversation history alone; a human-in-the-loop 
constraint under which outputs are presented as suggestions subject to user acceptance or 
rejection; and real-time operation coupled to the user's ongoing work. These characteristics 
position AI copilots as distinct from chatbots, which operate reactively within bounded 
conversational interfaces, and from autonomous AI agents, which execute multi-step goals with 
minimal human intervention. Representative deployments include GitHub Copilot for software 
development, Claude Code for agentic coding assistance, Microsoft 365 Copilot for enterprise 
productivity, and PathChat for clinical pathology~\cite{lu2024multimodal}, each reflecting a 
different instantiation of this paradigm across professional domains.

\subsection{Comparative Positioning}

\begin{figure*}[!t]
\centering
\includegraphics[width=0.80\textwidth]{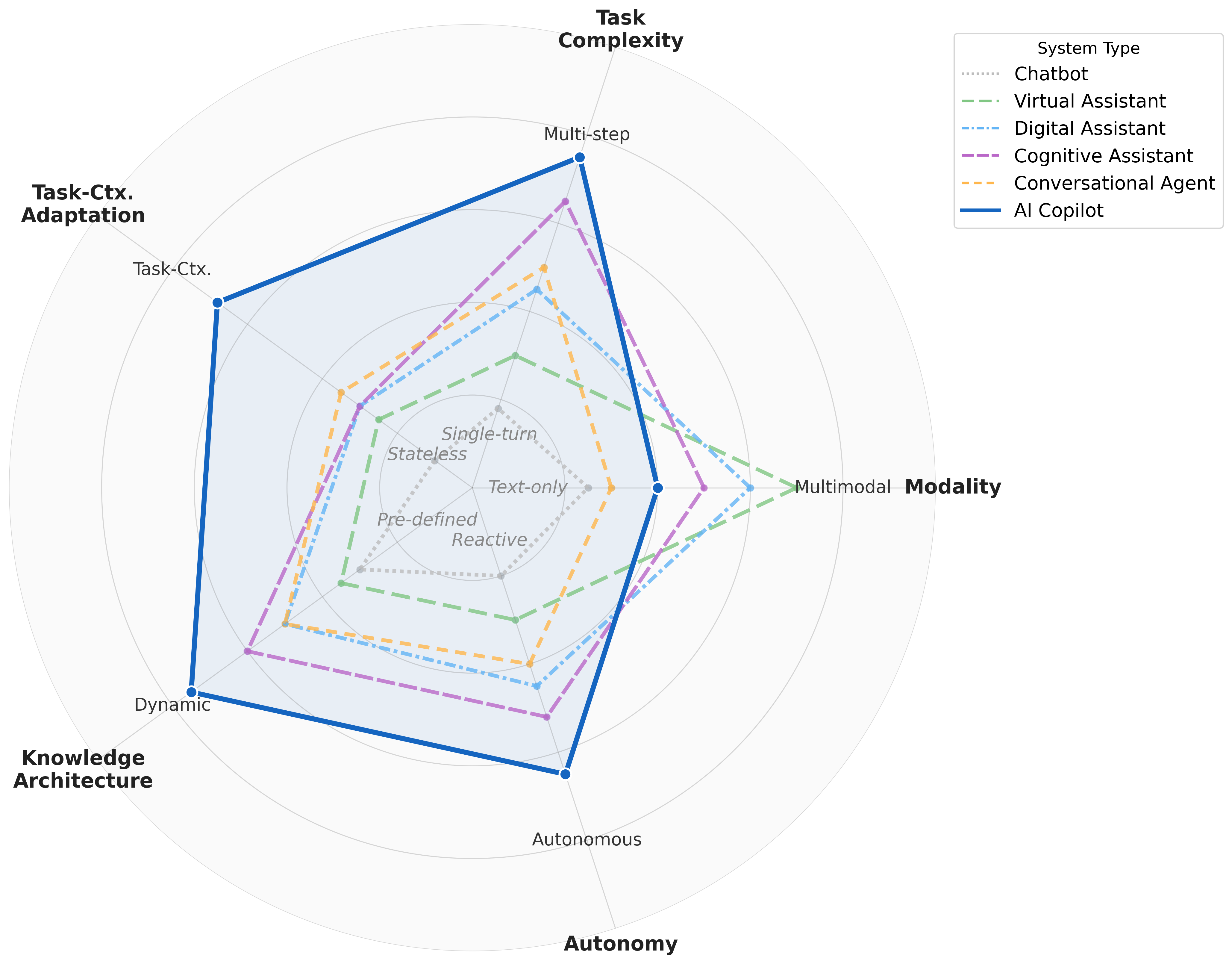}
\caption{Comparative positioning of AI assistive system types across five capability axes. Values represent relative positioning rather than absolute performance metrics. Task-Ctx. denotes Task-Contextualized Adaptation.}

\label{fig:assistant-comparison}
\end{figure*}

Fig.~\ref{fig:assistant-comparison} positions AI copilots against the assistive systems 
reviewed in Section~\ref{sec:background} across five axes: \textit{Modality}, \textit{Task 
Complexity}, \textit{Task-Contextualized Adaptation}, \textit{Knowledge Architecture}, 
and \textit{Autonomy}. The profiles reflect a fundamental architectural divide between 
systems designed for conversational accessibility and those designed for deep task 
integration.

Systems such as chatbots, virtual assistants, and digital assistants are optimized for 
broad accessibility across interaction modalities, a design priority that inherently 
limits task depth and preference specificity. Their stateless, session-bound nature 
represents a deliberate trade-off: generality is preserved at the cost of contextual 
grounding. Conversational and cognitive agents extend further into task complexity and 
knowledge reasoning, yet remain constrained by the absence of artifact-level context, 
through which fine-grained preference signals are captured.

AI copilots occupy a distinct position by embedding assistance directly within the 
user's workflow, where artifact context makes task-contextualized preference modeling 
possible. Deliberately bounded Autonomy is similarly a design commitment: preserving 
human oversight at the point of action is what enables continuous preference refinement, 
as user modifications to suggestions constitute the primary feedback signal. Notably, 
recent deployments such as Claude Code begin to relax this constraint by executing 
multi-step tasks autonomously while maintaining a reporting loop with the user, 
indicating that the boundary between AI copilot and autonomous agent remains an open 
design question. The mechanisms through which task-contextualized preference modeling 
is realized are examined in Section~\ref{sec:preference}.

%
%%%%%%%%%%%%%%%%%%%%%%%%%%%%%% SECTION END %%%%%%%%%%%%%%%%%%%%%%%%%%%%%%%%%

\section{Preference-based Response Generation}
\label{sec:preference}

\begin{figure*}[!t]
\centering
  \includegraphics[width=\textwidth]{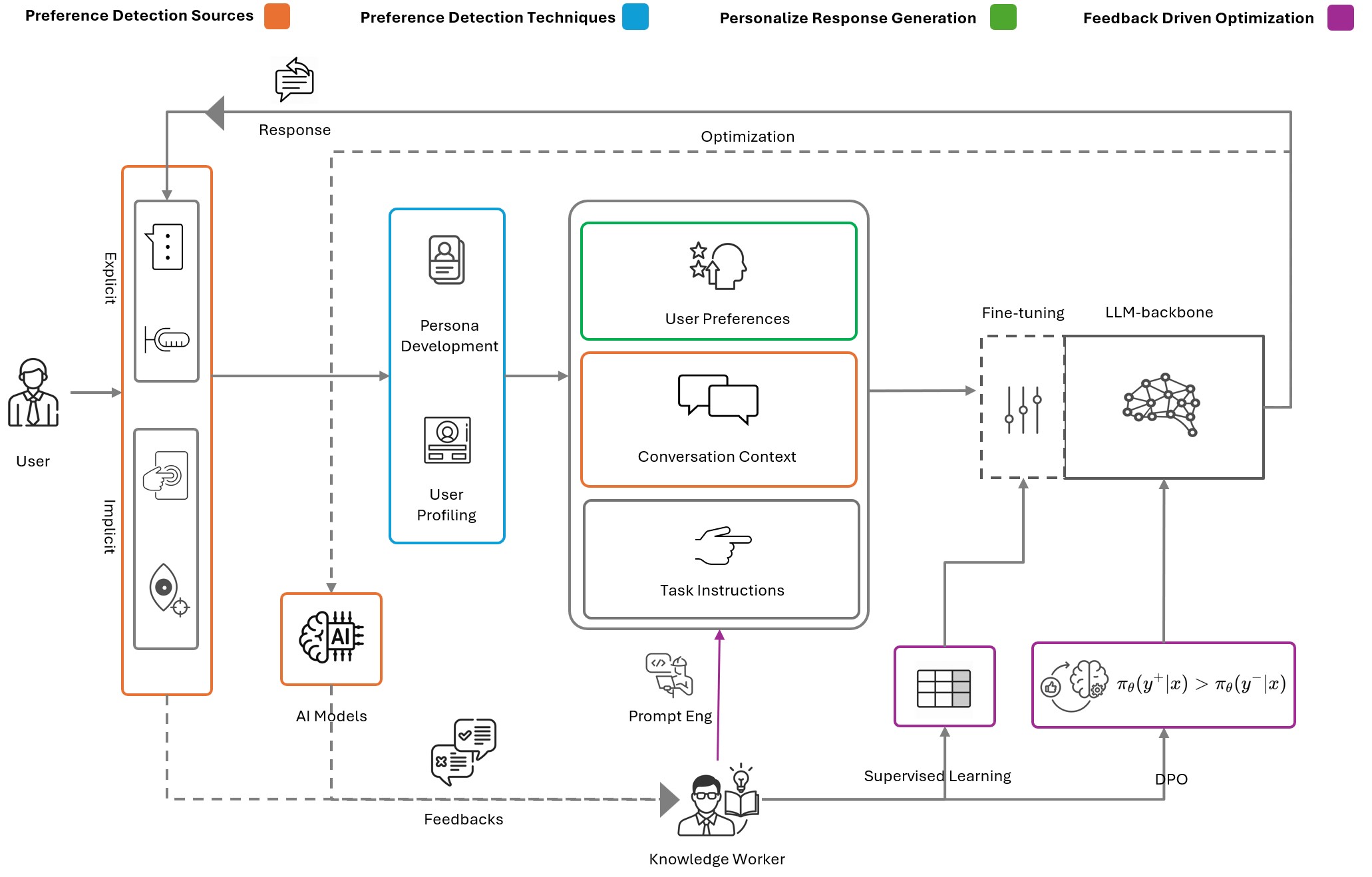}
  \caption{Conceptual architecture of a preference-aware AI copilot. The diagram illustrates the continuous feedback loop from user interaction and persona development to LLM-backed response generation and preference optimization. Knowledge Workers provide annotations, few-shot examples, and response pairs for supervised learning and DPO-based fine-tuning.}
  \label{fig:copilot-arch}
\end{figure*}

The ability to model and respond to user preferences must be treated as a foundational, system-level attribute of an AI copilot, necessitating rigorous integration across all stages of deployment. Fig.~\ref{fig:copilot-arch} maps the end-to-end architecture of a preference-aware AI copilot. The primary pathway illustrates personalized response generation: user interactions are captured through explicit and implicit channels, then processed to build user profiles and personas. These structured representations, combined with conversation context and task instructions, form a persona-aware prompt that conditions the LLM to generate personalized responses. A secondary pathway enables long-term adaptation through feedback-driven optimization. Here, knowledge workers process user feedback to provide annotations for supervised learning, few-shot examples for prompt engineering, or response pairs for preference optimization methods such as DPO. These signals update the LLM through fine-tuning, allowing the system to refine its behavior as user preferences evolve.

\subsection{Preference Detection Sources}

Personalized response generation relies on three related but distinct concepts that are often conflated in the literature. A \emph{preference signal} is raw, observable evidence of what a user wants, collected during or after interaction, such as pairwise rankings, binary feedback, or behavioral traces like clicks and edits. While preference signals capture what a user responds to in the moment, a \emph{user profile} takes a broader view, representing a structured, individual-level account of a specific user through factual attributes such as demographics, stated preferences, and interaction history. A \emph{persona} operates at a higher level of abstraction still, serving as a representation of a user group with similar behaviors and preferences \cite{afzoon2025exbigbang}, rather than describing any single individual. In practice, the available preference signals constrain which personalization techniques are feasible, as pairwise comparisons support preference learning from ranked outputs, while free-form edits provide lower-fidelity but low-friction supervision. Table~\ref{tab:preference_detection_sources} summarizes common sources of preference signals, highlighting how differences in signal fidelity, scalability, and supervision strength influence the choice of personalization and optimization strategies.

\begin{table*}[ht]
\centering
\caption{Preference detection sources, their typical use cases, 
and key limitations in AI copilots.}
\label{tab:preference_detection_sources}
\renewcommand{\arraystretch}{1.25}
\resizebox{0.85\textwidth}{!}{%

\begin{tabular}{l|l|l}
\hline
\textbf{Source Category} & \textbf{Typical Use Cases} & 
\textbf{Key Limitations} \\
\hline
\textbf{Explicit Feedback} \cite{zhao2018explicitimplicit, chiang2024chatbotarena, loepp2024generativeai} &
\makecell[l]{
Preference model training \\
Benchmarking and evaluation \\
High-fidelity supervision
} &
\makecell[l]{
High annotation cost \\
User fatigue \\
Limited coverage
} \\
\hline
\textbf{Implicit Human Feedback} \cite{mobasher2003webusage, heck2021gaze, tucker2024coactive} &
\makecell[l]{
Behavioral preference inference \\
Adaptive personalization \\
Continuous signal collection
} &
\makecell[l]{
Noisy observations \\
Privacy constraints \\
Ambiguous attribution
} \\
\hline
\textbf{Hybrid Feedback} \cite{park2024rlhf, cao2017userprofiling} &
\makecell[l]{
Long-term user modeling \\
Cold-start mitigation \\
Robust preference estimation
} &
\makecell[l]{
Signal inconsistency \\
Calibration complexity \\
Domain dependence
} \\
\hline
\textbf{Interactive Elicitation} \cite{zhang2024selfexploring, piriyakulkij2023active, handa2024bayesian, austin2024bayesian} &
\makecell[l]{
Sample-efficient preference learning \\
Ambiguity resolution \\
Cold-start personalization
} &
\makecell[l]{
User burden \\
Query design difficulty \\
Limited goal coverage
} \\
\hline
\textbf{LLM-Generated Signals} \cite{zhang2024gpg, go2023fdivergence, wu2025aloe, yoon2024evaluating} &
\makecell[l]{
Scalable supervision signals \\
User simulation \\
Long-context adaptation
} &
\makecell[l]{
Hallucination risk \\
Bias amplification \\
Requires grounding
} \\
\hline
\end{tabular}}
\end{table*}

\subsubsection{Human-Driven Feedback Channels}

Human-driven signals remain the most direct evidence of what a user prefers, though they differ substantially in fidelity, collection burden, and privacy risk. A useful organizing principle is to distinguish \emph{explicit} judgments, where users actively express preferences, from \emph{implicit} behavioral traces, where preferences are inferred from interaction patterns, and to treat \emph{interactive} and \emph{hybrid} methods as strategies that trade collection burden for information gain.

\paragraph{Implicit feedback.}
Implicit feedback is collected without asking the user to label 
preferences, making it low-friction but harder to interpret. Early 
work inferred preferences from click and navigation traces using 
web usage mining \cite{mobasher2003webusage}, while more recent 
work draws on richer cues such as gaze patterns as proxies for 
attention and interest \cite{heck2021gaze}. User editing behavior 
provides another applicable signal, where in coactive learning, 
small edits to model outputs are treated as weak preference 
feedback that can guide model updates \cite{tucker2024coactive}. 
The core limitation of implicit signals is attribution, as a 
click, dwell time, or edit may reflect convenience, curiosity, 
or interface effects rather than genuine preference. This 
ambiguity is compounded by privacy sensitivity, since collecting 
behavioral traces typically requires consent, data minimization, 
and careful storage practices. Heimdall addresses this by 
proposing privacy-respecting implicit preference collection 
through aggregation and anonymization mechanisms 
\cite{heimdall2017privacy}, though such safeguards introduce 
additional engineering complexity.

\paragraph{Explicit feedback.}
Explicit feedback directly asks users to evaluate outputs, 
offering higher signal fidelity than implicit traces at the cost 
of greater collection burden. It takes two main forms: pointwise 
labels such as ratings or satisfaction scores, and comparative 
judgments such as pairwise preference rankings. Pairwise 
comparisons are particularly common for evaluation benchmarking 
and preference model training, with Chatbot Arena being a 
prominent example that collects human preferences over model 
responses at scale \cite{chiang2024chatbotarena}. Choice-based 
elicitation can be augmented using generative interfaces to 
improve engagement and clarity during the elicitation process 
\cite{loepp2024generativeai}, while satisfaction-oriented labels 
have been shown to better capture user preference than 
engagement-only metrics in recommender settings 
\cite{zhao2018explicitimplicit}. The core limitation is 
annotation cost and user fatigue, since explicit feedback cannot 
be collected for every interaction at scale. A subtler limitation 
is coverage bias, as explicit signals are constrained by what 
the system chooses to ask about, leaving unqueried preference 
dimensions unrepresented. Together these constraints create a 
practical ceiling on how much explicit supervision any deployed 
system can realistically obtain.

\paragraph{Interactive and hybrid elicitation.}
Interactive elicitation addresses the cost limitations of explicit 
feedback by asking questions only when the system is uncertain, 
targeting the most informative preference queries rather than 
collecting feedback uniformly. Self-exploring language models 
exemplify this approach, generating targeted alternatives to probe 
user preference efficiently for online alignment 
\cite{zhang2024selfexploring}. Related methods use probabilistic 
modeling to represent uncertainty and update beliefs incrementally 
from language feedback \cite{piriyakulkij2023active, 
handa2024bayesian}, while Bayesian optimization further reduces 
interaction cost by selecting prompts that are expected to yield 
the greatest information gain \cite{austin2024bayesian}. The 
limitation is query design complexity, as poorly constructed 
queries can confuse users or fail to surface the preference 
dimensions that matter most.

Hybrid methods take a different approach to the same problem, 
combining explicit and implicit evidence to compensate for the 
sparsity or unreliability of either source alone. Heterogeneous 
feedback aggregation studies how to fuse different signal types 
in RLHF-style settings \cite{park2024rlhf}, while topical 
profiling integrates explicit self-declarations with implicit 
behavioral footprints such as hashtag usage and interaction logs 
\cite{cao2017userprofiling}. The practical challenge is 
engineering complexity, since signals must be normalized, 
conflicts resolved, and privacy safeguards enforced across 
sources that may differ substantially in format and reliability.

Together, these human-driven channels provide high-confidence 
preference evidence but vary substantially in scalability, 
motivating the complementary use of model-generated signals 
reviewed in the following subsection.

\subsubsection{LLM-Generated Feedback Channels}

The reliance of human-driven channels on manual labeling has 
motivated a parallel line of work that uses LLMs to produce 
auxiliary preference signals for personalization and evaluation. 
While these signals reduce annotation cost and scale more readily 
than human feedback, they introduce a distinct category of risk, 
as LLM-generated signals can encode incorrect, outdated, or 
hallucinated preferences that propagate silently through the 
personalization pipeline.

\paragraph{Profile synthesis.}
Rather than conditioning generation directly on raw interaction 
context, some methods synthesize a structured user profile as an 
intermediate representation. In Guided Profile Generation, an LLM 
produces an intermediate profile that improves downstream 
personalization by increasing controllability and modularity 
\cite{zhang2024gpg}. The limitation is grounding, as the quality 
of the profile depends entirely on the reliability of the evidence 
it is derived from. A confidently generated but inaccurate profile 
can steer personalization systematically in the wrong direction, 
making this approach sensitive to the quality of the underlying 
interaction data.

\paragraph{Objective-driven alignment.}
LLM-generated signals can also be incorporated through explicit 
optimization objectives rather than intermediate representations. 
Aligning outputs via f-divergence minimization provides a 
principled way to control shifts toward preferred responses while 
balancing stability and diversity \cite{go2023fdivergence}. 
However, these objectives are only as reliable as the preference 
evidence they optimize against, and noisy or biased signals can 
produce brittle alignment that degrades under distribution shift.

\paragraph{Interaction-driven adaptation and user simulation.}
Beyond single-turn signal generation, LLM-generated feedback can 
support long-context personalization adaptation through multi-turn 
interaction. An interaction-based framework for aligning LLMs with 
individual preferences demonstrates that repeated exchanges 
progressively improve alignment quality \cite{wu2025aloe}. 
Separately, LLMs can serve as generative user simulators, enabling 
controlled evaluation and systematic stress testing of 
conversational recommender behavior at scale \cite{yoon2024evaluating}. 
Simulation is particularly valuable for cold-start settings where 
real user data is scarce, though simulated behavior may diverge 
from genuine user responses in ways that are difficult to detect 
without periodic human validation. Across all three strategies, 
LLM-generated signals are best treated as complements to human 
supervision rather than replacements, requiring grounding checks 
and robust benchmarks to remain trustworthy.

%%%%%%%%%%%%%%%%%%%%%%%%%%%%%%%%%%%%%%%%%%%%%%%%
\subsection{Preference Detection Techniques}

Following the identification of diverse preference signal sources, the architectural focus shifts to the algorithmic techniques utilized to detect, extract, and model these preferences across the interaction lifecycle. While the techniques discussed in this section have been primarily developed within the broader context of recommender systems and conversational agents, their fundamental mechanisms are directly applicable to the architecture of AI copilots, enabling them to align with user intents dynamically \cite{zhou2022user}.

%% NEW CONTENT:
\subsubsection{Persona Representations Taxonomy}

A prerequisite for selecting any preference detection technique is determining how the user will be represented internally, the persona model. This choice is consequential: the structural form of the persona directly constrains which detection algorithms and personalization techniques are applicable downstream. A system cannot apply gradient-based alignment to a static declarative profile, nor can it perform symbolic relation extraction on a continuous embedding vector \cite{zhou2022user, eke2019survey}. Four persona representation paradigms are identified here, each associated with a distinct structural form, interaction phase, and set of compatible techniques, as summarized in Table~\ref{tab:persona_taxonomy}.

\textbf{Explicit statements} are declarative, rule-based profiles constructed from user-reported attributes such as demographics, stated preferences, or role descriptions. Their structured, human-readable form makes them directly injectable into prompt templates and retrieval pipelines before any interaction begins \cite{eke2019survey}. The limitation is coverage: explicit statements capture only what users are willing and able to articulate, leaving implicit preferences unrepresented.

\textbf{Interaction trajectories} encode the user as a temporal sequence of behavioral events, clickstreams, navigation logs, or historical dialogue turns. This form supports both pre- and mid-interaction personalization, as trajectories can be collected before a session and updated continuously as it unfolds. Sequence-aware models and transformer architectures are the natural fit, designed to capture temporal dependencies across variable-length histories \cite{choi2021user, he2017neural}. The tradeoff is signal reliability: behavioral traces are noisy proxies for preference, as the same action can reflect habit or interface constraints rather than genuine intent.

\textbf{Personal knowledge bases} represent the user through structured, symbolic stores, knowledge graphs or vector databases, populated from conversational content in real time. This paradigm is most applicable during active interaction, where utterances can be parsed into relation triplets and stored as auditable preference parameters \cite{zhu2023PAED, lim2023beyond}. Retrieval-augmented generation pipelines can then query these stores to condition responses on structured user knowledge. The constraint is rigidity: symbolic schemas struggle with ambiguous or implicit preferences that do not map cleanly to predefined relations.

\textbf{Latent embeddings} represent the user as continuous vector representations shaped by optimization over feedback signals. This is the domain of post-interaction alignment methods, RLHF, DPO, and Kahneman-Tversky Optimization (KTO), which update model parameters directly from preference evidence \cite{ouyang2022training, rafailov2023direct, ethayarajh2024kto}. Latent representations are the most expressive of the four paradigms, capable of encoding subtle preferences that no explicit schema could capture. Their limitation is opacity: the persona exists only implicitly in model weights, making it difficult 
to inspect or correct when alignment fails.

These four paradigms form a design space rather than a progression. The choice of representation, or combination of representations, depends on what feedback signals are available, which interaction phase the system operates in, and what tradeoff between expressiveness and interpretability is acceptable for the deployment context.

\begin{table*}[!t]
\centering
\caption{Persona representation paradigms and their compatible 
personalization techniques, mapped across interaction phases. 
The choice of representation constrains which techniques are 
applicable at each stage.}
\renewcommand{\arraystretch}{1.3}
\resizebox{0.85\textwidth}{!}{%
\begin{tabular}{c|c|c}
\hline
\textbf{Representation Paradigm} & 
\textbf{Primary Interaction Phase} & 
\textbf{Compatible Techniques} \\
\hline
Explicit Statements & 
Pre-Interaction & 
\makecell[l]{
Prompt-engineering \\
Retrieval pipelines \cite{eke2019survey}
} \\
\hline
Interaction Trajectories & 
Pre- \& Mid-Interaction & 
\makecell[l]{
Sequence-aware models \\
Transformers \cite{choi2021user, he2017neural}
} \\
\hline
Personal Knowledge Bases & 
Mid-Interaction & 
\makecell[l]{
RAG \\
Relation triplet extractors \cite{zhu2023PAED, lim2023beyond}
} \\
\hline
Latent Embeddings & 
Post-Interaction & 
\makecell[l]{
RLHF, DPO, KTO \\
Gradient-based alignment \cite{ouyang2022training, 
rafailov2023direct, ethayarajh2024kto}
} \\
\hline
\end{tabular}
}
\label{tab:persona_taxonomy}
\end{table*}

\begin{figure*}[!t]
\centering
  \includegraphics[width=0.95\textwidth]{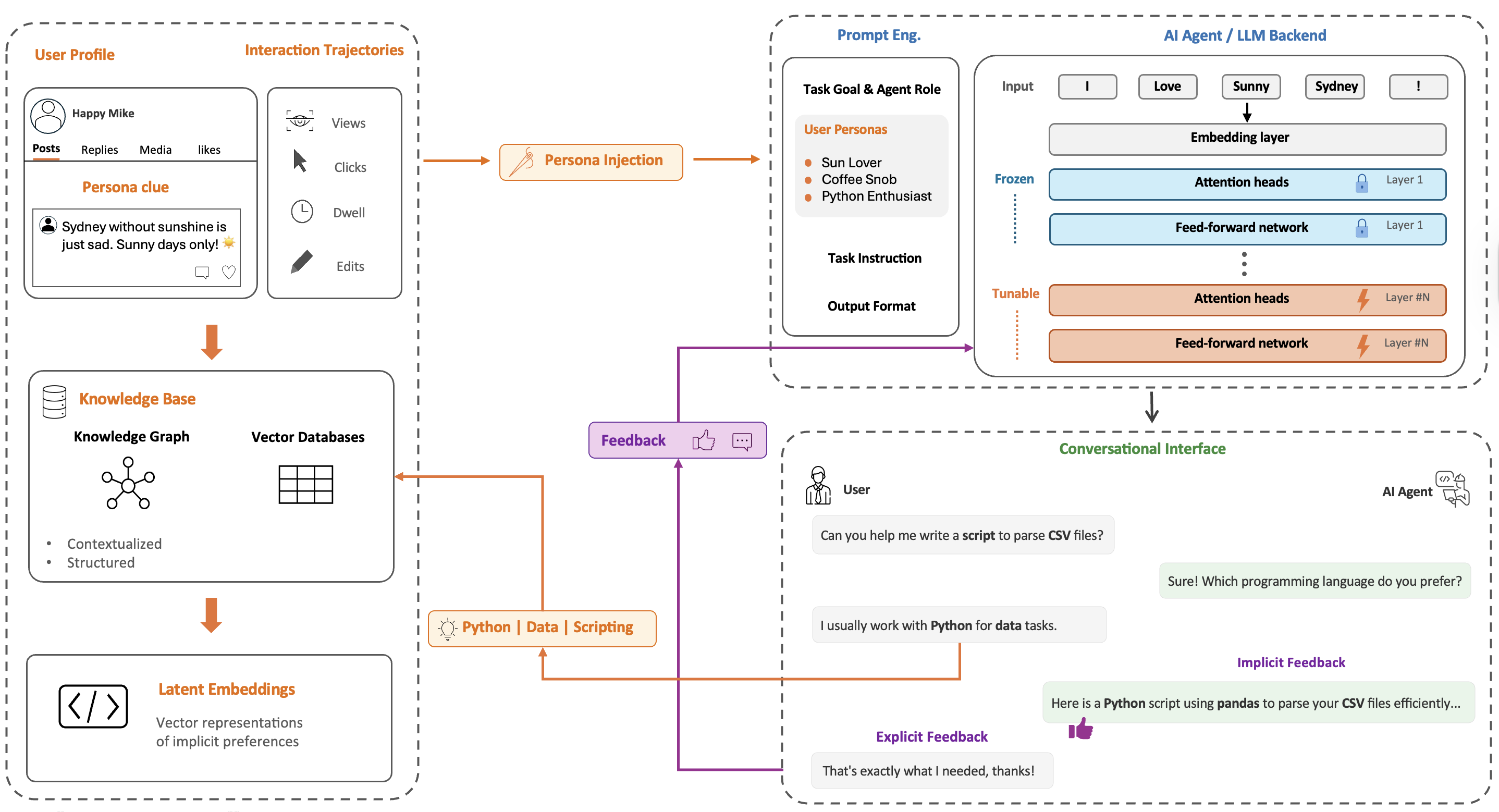}
\caption{Integration of persona representations into AI copilot systems. Persona information from user profiles, interaction trajectories, knowledge bases, and latent embeddings is injected via prompt engineering to condition the LLM, with user feedback enabling continuous refinement.}

\label{fig:conversation_phases}
\end{figure*}

By mapping these distinct persona representations across the pre-interaction, mid-interaction, and post-interaction phases, conversational agents can dynamically transition from coarse-grained static profiling to highly nuanced, continuous behavioral alignment \cite{zhou2022user}. Fig.~\ref{fig:conversation_phases} provides a detailed temporal visualization of this extraction mechanism. In the pre-interaction phase, the system relies on predefined user profiles and historical interaction trajectories to mitigate the cold-start problem, establishing a baseline persona. As the session transitions into the mid-interaction phase, the AI copilot leverages real-time conversational context and workspace artifacts to dynamically populate personal knowledge bases and vector databases. Finally, in the post-interaction phase, the accumulated explicit and implicit feedback is transformed into latent embeddings. These continuous vector representations are then utilized by gradient-based alignment algorithms to permanently shift the model's parametric weights, ensuring that the persona evolves in tandem with the user's changing preferences.

%%%%%%%%%%%%%%%%%%%%%%%%%%% 

\subsubsection{Cold-Start Persona Modeling (Pre-Interaction)} 

Before interaction begins, systems must rely on predefined profiles, historical population data, or early preference elicitation to address the cold-start problem. Gaussian Processes (GPs) were among the first approaches applied to this phase, used for Bayesian preference elicitation because they avoid the restrictive assumption of parametric linear utility functions, modeling latent user utility directly over the joint space of user and item features \cite{chu2005preference}. A key advantage is the capacity to quantify predictive uncertainty: by maintaining a distribution over possible utility functions, the system can apply Value of Information heuristics to select the most discriminative preference questions, reducing elicitation burden for new users \cite{chu2005preference}. However, standard GP inference requires matrix inversion that scales cubically ($\mathcal{O}(N^3)$) with the number of data points, making GP-based preference models computationally impractical for the large concurrent user bases typical of modern deployments \cite{chu2005preference}.

To address this scalability constraint, the field shifted toward Collaborative Filtering (CF) and its deep learning extensions. Early CF models used matrix factorization, representing users and items as latent vectors combined via inner products \cite{he2017neural}. Neural Collaborative Filtering (NCF) replaced this fixed inner product with a multi-layer perceptron, enabling the network to learn non-linear interaction functions from historical data \cite{he2017neural}. Social NCF variants incorporate implicit feedback, such as clickstreams, dwell times, and document views, which require no active user participation \cite{he2017neural}. This reliance on implicit signals introduces a brittle assumption: unobserved interactions are treated as negative preferences, when in reality they may reflect missing data or undiscovered content. The result is a scarcity of genuine negative feedback that makes the model vulnerable to popularity bias, conflating global trends with individual preference \cite{he2017neural}. Deeper NCF architectures improve representational capacity but reduce interpretability and increase overfitting risk on sparse user interaction graphs \cite{he2017neural}.

To incorporate semantic context that interaction-based CF models cannot capture, sequence-aware models such as SessionBERT extract personas from behavioral trajectories \cite{choi2021user}. A user's sequential navigation, reading metadata, clicking action buttons, moving through interface states, is treated as a semantic language of intent, with a transformer trained using a masked language modeling objective on discrete trajectory tokens \cite{choi2021user}. This captures contextual and temporal dependencies that NCF cannot represent, enabling zero-shot persona identification before the user provides any explicit input \cite{choi2021user}.

%%%%%%%%%%%%%%%%%%%%%%%%%%%%

\subsubsection{Dynamic Preference Adaptation (Mid-Interaction)}

Once an active session begins, static profiles become unreliable when the user's immediate context diverges from historical patterns. Real-time persona extraction addresses this by shifting from profile retrieval to dynamic in-context preference inference \cite{cheng2024InDialog}.

Zero-Shot Persona Attribute Extraction in Dialogues (PAED) frames this as a relation triplet extraction task, identifying subjects, relations, and objects directly from conversational utterances, for example, extracting \textit{User - prefers framework - PyTorch} \cite{zhu2023PAED}. To overcome the absence of domain-specific annotated training data, PAED uses a contrastive learning architecture with hard negative sampling, enabling real-time population of a structured symbolic knowledge base from dialogue \cite{zhu2023PAED}. The resulting preference parameters are interpretable and directly usable in retrieval-augmented generation prompts \cite{zhu2023PAED, huang2024learning}. The limitation is rigidity: PAED assumes preferences map cleanly to predefined relational schemas, and accuracy degrades on ambiguous grammar, implicit context, and negations expressed through sarcasm or comparison \cite{zhu2023PAED}.

To address this rigidity, In-Dialogue Learning (IDL) takes a generative approach, fine-tuning the LLM to characterize the persona recursively from evolving dialogue history alone, without predefined profiles or relational slots \cite{cheng2024InDialog}. IDL operates through two stages, Mutual Supervised Learning and Deep Personalized Alignment, using Static and Dynamic Persona Identification to cluster and reorder historical turns, minimizing semantic disruption via conversational edit distance metrics \cite{cheng2024InDialog}. This allows incremental updates to an abstract user representation without structural constraints. The tradeoff is stability: because the persona is maintained within the context window or through soft-prompt updates, prolonged conversations or rapid context switches can cause the preference profile to degrade \cite{cheng2024InDialog}. Processing extensive dialogue histories also increases inference latency and memory overhead during sustained assistance tasks \cite{cheng2024InDialog}.

\subsubsection{Continuous Preference Refinement (Post-Interaction)} \label{sec:persona_detection}

Post-interaction feedback mechanisms are responsible for the long-term alignment of the system's foundational policy. By analyzing completed interaction outcomes, these methods update model parameters to prevent future misalignment, shifting the system from reactive response generation toward continuous behavioral adaptation \cite{ouyang2022training}.

The choice of feedback modality directly affects pipeline stability, scalability, and annotation cost \cite{rafailov2023direct}. RLHF using Proximal Policy Optimization (PPO) remains the foundational approach, but its reinforcement learning optimization is prone to instability and reward hacking, requiring careful reward model design and hyperparameter tuning \cite{ouyang2022training, casper2023open}. DPO addresses this by bypassing explicit reward modeling and optimizing directly on pairwise preference data, a more stable alternative whose design and variants are discussed in detail in Section~\ref{sec:dpo}. While more stable than PPO-based RLHF, DPO remains sensitive to preference noise and length bias \cite{rafailov2023direct}.
To address the data collection burden of pairwise methods, KTO learns from pointwise binary feedback, a single positive or negative signal per response, grounded in Kahneman-Tversky prospect theory \cite{ethayarajh2024kto}. Its asymmetric log-sigmoid loss prevents vanishing gradients when implicit rewards become strongly negative, enabling stable updates even on sparse feedback \cite{ethayarajh2024kto}. This makes KTO well-suited for production environments where pairwise annotation is impractical and binary telemetry can be collected at scale.

When high-quality pairwise data is available, DPOC provides additional robustness against preference misranking, the case where both candidate responses are poor but one is marginally less so \cite{cheng2024dialogues}. A criterion-based penalty ensures the preferred response surpasses an intermediate-quality reference baseline, preventing the model from optimizing toward relatively better but objectively inadequate outputs 
\cite{cheng2024dialogues}.

Beyond isolated feedback loops, AFSPP models long-term persona evolution by simulating perception and social influence within a closed environment, processing post-interaction outcomes to shift model weightings autonomously \cite{weng2023afspp}. Experimental results map agent preferences to psychological constructs such as the Big Five traits, though this rests on the assumption that such constructs reliably correspond to LLM latent spaces, introducing both computational overhead and the risk of emergent behaviors incompatible with deployment constraints \cite{weng2023afspp}.
To reduce the inefficiency of random exploration in continuous learning environments, Active Preference Learning frameworks use Bayesian inference to manage when feedback is solicited, triggering queries only when output entropy exceeds a meaningful uncertainty threshold \cite{lee2023active}. This maximizes informational value per interaction while limiting user feedback fatigue \cite{lee2023active}.

\subsubsection{Critical Synthesis of Preference Trade-offs}

The selection of preference detection architectures requires navigating distinct computational and operational trade-offs across interaction phases. In the \textbf{Pre-Interaction} phase, Gaussian Processes offer strong uncertainty quantification and sample efficiency, making them well-suited for high-stakes settings where active elicitation is acceptable, but their non-parametric inference relies on matrix inversion that scales cubically, making them impractical at deployment scale \cite{chu2005preference}. NCF frameworks scale to large user bases but rest on the assumption that unobserved interactions signal negative preference, which introduces popularity bias when that assumption breaks down \cite{he2017neural}. SessionBERT addresses this by treating sequential user actions as a semantic language of intent, capturing contextual dependencies that NCF cannot, at the cost of higher tokenization and training overhead \cite{choi2021user}.

During the \textbf{Mid-Interaction} phase, the core trade-off is between structural consistency and generative flexibility. PAED assumes preferences can be mapped to predefined relational schemas, which makes its output interpretable and auditable but causes accuracy to degrade on ambiguous or implicit input \cite{zhu2023PAED}. IDL removes this structural constraint by characterizing the persona dynamically from dialogue history alone, but because the persona is maintained within the context window, prolonged or shifting conversations introduce persona drift risk \cite{cheng2024InDialog}.

In the \textbf{Post-Interaction} phase, feedback modality drives the key trade-offs. KTO learns from pointwise binary signals, avoiding the paired annotation requirement of DPO and making it better suited for continuous production environments, though pointwise feedback provides a weaker supervision signal than pairwise comparisons, which can limit reasoning capability on complex preference distinctions \cite{ethayarajh2024kto}. DPOC improves resistance to reward hacking by requiring the preferred response to surpass a reference baseline, but this requires curated baseline data that may not always be available \cite{cheng2024dialogues}. AFSPP enables autonomous persona evolution over time but rests on the experimental assumption that psychological constructs map reliably to LLM latent spaces, introducing both computational overhead and unpredictability risk \cite{weng2023afspp}. Across all phases, Active Preference Learning helps manage data acquisition efficiency and limit user feedback fatigue by triggering queries only at points of meaningful model uncertainty \cite{lee2023active}. Table~\ref{tab:preference_detection} summarizes these methods and their key trade-offs.

%% STRATEGY: The Mechanism/Key Assumption column is removed — its content 
%% is moved into the Critical Synthesis prose where it strengthens the 
%% tradeoff analysis. Advantages and Limitations cells are reduced to 
%% short keywords. The result is a clean reference table that summarizes 
%% methods without duplicating prose content.

%% NEW CONTENT:
\begin{table*}[!t]
\centering
\caption{Preference detection methods across interaction phases, 
compared by key strength and limitation.}
\renewcommand{\arraystretch}{1.3}
\resizebox{0.85\textwidth}{!}{%
\begin{tabular}{c|l|l|l}
\hline
\textbf{Phase} & \textbf{Method} & \textbf{Key Strength} & 
\textbf{Key Limitation} \\
\hline
\multirow{3}{*}{Pre-Interaction} & 
GPs \cite{chu2005preference} & 
Uncertainty quantification & 
$\mathcal{O}(N^3)$ scaling \\
& SNCF \cite{he2017neural} & 
Scales to large user bases & 
Popularity bias \\
& SessionBERT \cite{choi2021user} & 
Semantic trajectory modeling & 
Computational overhead \\
\hline
\multirow{2}{*}{Mid-Interaction} & 
PAED \cite{zhu2023PAED} & 
Interpretable, auditable output & 
Fails on ambiguity \\
& IDL \cite{cheng2024InDialog} & 
No predefined schema needed & 
Persona drift risk \\
\hline
\multirow{3}{*}{Post-Interaction} & 
KTO \cite{ethayarajh2024kto} & 
Pointwise feedback, scalable & 
Lower reasoning capability \\
& DPOC \cite{cheng2024dialogues} & 
Reward hacking resistance & 
Requires curated baselines \\
& AFSPP \cite{weng2023afspp} & 
Autonomous persona evolution & 
Experimental assumptions \\
\hline
\end{tabular}
}
\label{tab:preference_detection}
\end{table*}

%%%%%%%%%%%%%%%%%%%%%%%%
\subsection{Personalized Response Generation}

Considering that the usable clues for preference optimization are now available through any of the previously discussed techniques, this section reviews different groups of approaches for response personalization. To this end, we have selected the techniques based on their technical diversity, rather than following a progression-based perspective on how this context has advanced over time. Table~\ref{tab:personalized_response_table} presents a high-level comparison of these response personalization studies applied at various levels of dialogue system generation, detailing their key advantages and limitations. In reviewing these studies, we considered a shared set of personalization aspects, such as consistency, coherence, and diversity, as identified in previous work \cite{afzoon2024persobench, chen2024recent}.

\begin{table*}[!t]
\centering
\caption{Comparison of personalized response generation approaches based on their advantages and limitations.}
\renewcommand{\arraystretch}{1.3}
\resizebox{0.85\textwidth}{!}{%
\begin{tabular}{l|l|l}
\hline
\textbf{Personalization Level} & \textbf{Key Advantages} & \textbf{Key Limitations} \\
\hline
\textbf{Prompt-Based} \cite{dong2025cross, liu2024context, kim2020will, oh2023pk, huang2024selective} & \makecell[l]{ Model-agnostic \\ Low computational cost \\ Flexible to context } & \makecell[l]{ Shallow persona integration \\ Dependent on retrieval quality } \\
\hline
\textbf{Fine-Tuning-Based} \cite{wolf2019transfertransfo, wang2020cue, cao2022model, han2023personapkt, varma2024talk, tan2024personalized} & \makecell[l]{ Memory-efficient \\ Fast adaptation \\ Low storage overhead } & \makecell[l]{ Limited deep reasoning \\ Risk of overfitting on sparse persona data } \\
\hline
\textbf{Architecture-Level} \cite{bao2019plato, wu2021personalized, ju2022learning, mahajan2024persona, zhong2020towards, huang2023personalized, lim2023beyond} & \makecell[l]{ High expressiveness \\ Fine-grained structural reasoning } & \makecell[l]{ High training cost \\ Requires large labeled datasets } \\
\hline
\end{tabular}%
}
\label{tab:personalized_response_table}
\end{table*}

\subsubsection{Prompt-Based Personalization}
As a straightforward class of methods, prompt-based approaches steer the model's behavior by modifying the input through prompt engineering or retrieval-augmented generation (RAG), rather than altering the model itself. Their key advantages include model-agnostic applicability, low computational cost, and flexibility at inference time, making them well-suited for lightweight or rapidly deployable systems focused on achieving persona consistency, persona diversity, persona-grounded knowledge, and dynamic style adaptation.

A variety of persona coverage and injection approaches have been proposed to carefully construct inputs without altering the underlying language model. To address the challenge of encoding both user-specific and conversational context, Cross-Graph Knowledge Exchange (CKE) \cite{dong2025cross} constructs a hybrid structured prompt by first generating dialogue user graphs for both conversation participants and then performing cross-graph knowledge aggregation. Discrete and continuous representations are fused to form a prompt that encodes persona, dialogue, and concept-level relationships for input to a frozen LLM. Similarly, to handle domain-specific personalization where relevant context is scattered across long dialogue histories, a topic-focused summarization method \cite{liu2024context} distills relevant context segments from the dialogue history. These summaries, combined with structured profile attributes and topic-aligned dialogue segments, are assembled into a prompt to guide generation in domain-specific tasks.

Beyond structural augmentation, token-level and retrieval-based approaches offer complementary mechanisms for finer-grained control. To address the problem of generating responses that are distinctive to a given persona rather than generic, Kim et al. \cite{kim2020will} employ token-level probabilistic reasoning using the Rational Speech Acts (RSA) framework. The core mechanism models a self-conscious speaker that adjusts token probabilities by simulating how an imaginary listener would infer the speaker's persona from the utterance. A distractor memory storing alternative persona variants supports this inference, enforcing utterance distinctiveness and consistency during decoding. The limitation is computational overhead, as the RSA framework requires multiple forward passes to compute pragmatic adjustments. To address the complementary problem of grounding responses in external knowledge while maintaining persona consistency, PK-ICR \cite{oh2023pk} implements a dual retrieval prompting strategy that leverages both dialogue context and persona to retrieve knowledge facts through a neural QA retriever, enabling precise grounding for generation. Finally, to address the rigidity of fixed prompts, Selective Prompt Tuning (SPT) \cite{huang2024selective} introduces a soft prompt selection mechanism, where multiple trainable soft prompts are maintained and a dense retriever selects the most suitable prompt per input. Contrastive and fusion learning objectives are applied to ensure prompt diversity and stability across dialogue contexts.

\subsubsection{Fine-Tuning-Based Personalization}
While prompt-based methods are simple and model-agnostic, they offer limited control over model behavior because persona information remains external to the model's parameters. In contrast, fine-tuning-based personalization enables more robust and persistent behavioral adaptation by updating model weights, making it suitable when deep style adaptation, emotion alignment, naturalness, fluency, and interpretability are required beyond what input conditioning alone can achieve.

To address the challenge of building persona-consistent dialogue agents without task-specific architectures, TransferTransfo \cite{wolf2019transfertransfo} demonstrated that transfer learning from large pretrained language models could achieve state-of-the-art performance on persona-based dialogue. The approach fine-tunes a pretrained GPT model using a multi-task objective that combines language modeling loss with next-utterance classification. The input representation concatenates persona sentences, dialogue history, and the response candidate with special separator tokens, allowing the model to learn persona-sensitive representations without architectural modification. This approach achieved a 45\% improvement in perplexity on the ConvAI2 PersonaChat benchmark, establishing transfer learning as the dominant paradigm for persona-based dialogue. The limitation is that persona consistency depends entirely on the attention mechanism learning to attend to relevant persona sentences, which can fail when personas are long or contain conflicting information.

To address the problem of dynamically weighting different conditioning signals during generation, Cue Adaptive Decoder (CueAD) \cite{wang2020cue} introduced a decoder-side mechanism using GRU+, a modified GRU with a gating function that selectively incorporates conditioning cues, such as persona or dialogue context, based on their relevance at each decoding step. This enabled dynamic and interpretable control over cue influence, though at the cost of requiring cue-specific architectural modifications. Taking a different approach focused on data rather than architecture, the D³ framework \cite{cao2022model} applies a three-stage fine-tuning pipeline: distillation to ensure alignment with a teacher model, diversification to expand behavioral coverage through data augmentation, and curriculum training to manage input complexity. This structured process improved generalization and consistency without requiring architectural changes.

Parameter-efficient methods have since emerged to reduce computational overhead while preserving personalization capability. To address the storage cost of maintaining separate fine-tuned models for each user, PersonaPKT \cite{han2023personapkt} adopts prefix tuning, where continuous prefix vectors are injected into a frozen backbone. A shared prefix is first trained across multi-persona data and then specialized into persona-specific prefixes, enabling private, modular adaptation with minimal parameter overhead. Extending parameter efficiency to emotional adaptation, Varma et al. \cite{varma2024talk} applied LoRA-based fine-tuning to an LLaMA-3 model, where an emotion classifier triggers prompt selection to guide responses conditioned on emotional tone and demographic context. Building on this direction, Tan et al. \cite{tan2024personalized} proposed a method in which small, LoRA-based modules, referred to as pieces, are retrieved and composed at inference time based on user characteristics. These modules enable storage-efficient, task-free personalization without altering the base model, though the retrieval step introduces latency at inference time.

\subsubsection{Architecture-Level Personalization}
Despite the flexibility offered by fine-tuning methods, they remain bounded by the representational capacity of the underlying architecture. When personalization demands richer structural modeling, such as achieving deep persona grounding, controlling dialogue acts, maintaining persona coherency, or explicitly handling multi-party dialogue setups, modifications to the model's architecture become necessary. At this level of intervention, personalization is achieved by introducing new components or redesigning internal mechanisms to explicitly encode and integrate user-specific signals.
 
A fundamental challenge in dialogue generation is the one-to-many problem: a single context admits multiple valid responses with different intents, tones, or topics. To address this, PLATO \cite{bao2019plato} introduced discrete latent variables into the transformer architecture, where each latent value corresponds to a latent speech act representing a specific dialogue intent. The model is trained with two reciprocal tasks performed simultaneously within a shared network: response generation, which maximizes the likelihood of the target response given the context and latent variable, and latent act recognition, which estimates the latent variable given both context and target response. This dual-task training allows the model to learn diverse response patterns without requiring explicit dialogue act annotations. At inference time, multiple responses can be generated by sampling different latent values, enabling controllable diversity. The approach achieved strong results on PersonaChat and knowledge-grounded dialogue benchmarks, though the discrete latent space can limit expressiveness compared to continuous alternatives.

Building on the need to capture different dimensions of user information, memory-based architectures were proposed to separate static persona attributes from dynamic interaction patterns. Wu et al. \cite{wu2021personalized} encode user profiles and interaction histories into split memory networks, where multi-hop attention mechanisms allow selective retrieval of static and dynamic user signals during generation. This separation enables the model to ground responses in stable persona traits while remaining responsive to conversational context.
 
To better model interactions in multi-party settings where multiple speakers with distinct personas participate, graph-based structures have been introduced to capture the relational structure of conversations. Ju et al. \cite{ju2022learning} encode utterances, speakers, and dialogue flow into a heterogeneous graph, incorporating an addressee prediction module trained with posterior supervision to model who is speaking to whom. This line of work was extended by Mahajan and Shaikh \cite{mahajan2024persona}, who incorporated user persona attributes directly as graph nodes connected via meta-relational edges, enabling fine-grained reasoning about persona-persona and persona-utterance relationships.

Complementary approaches have focused on the response selection rather than generation setting. CoBERT \cite{zhong2020towards} leverages multi-hop co-attention to capture second-order interactions between context, persona, and candidate responses, supporting persona-consistent selection in retrieval-based setups. For generative models, adaptive attention mechanisms have been introduced \cite{huang2023personalized} where cross-attention over persona and dialogue context is modulated through dynamic weighting and masking to filter irrelevant signals at each decoding step. Finally, to address the challenge of sparse or fragmented persona inputs, broader architectural pipelines have been designed that combine concept-based persona generation with alignment modules before feeding curated persona content into the generation process \cite{lim2023beyond}.

\subsubsection{Critical Synthesis}
 
The choice among prompt-based, fine-tuning-based, and architecture-level personalization involves trade-offs across computational cost, deployment flexibility, and personalization depth.
 
\textbf{Prompt-based methods} offer the lowest barrier to deployment: they require no training, work with any frozen LLM, and can be updated instantly by modifying the prompt. However, persona integration remains shallow because the model has no learned parameters dedicated to personalization. These methods are best suited for rapid prototyping, deployment with closed-source models, or settings where personas change frequently and retraining is impractical.
 
\textbf{Fine-tuning-based methods} offer deeper personalization by encoding persona-relevant patterns into model weights, improving consistency and naturalness compared to prompt-only approaches. Parameter-efficient variants such as prefix tuning and LoRA reduce storage costs to a few megabytes per persona, making multi-user deployment feasible. The trade-off is that fine-tuning requires labeled persona-dialogue data, and models can overfit when such data is sparse. These methods are best suited for production systems with stable user bases where persona data can be collected over time.
 
\textbf{Architecture-level methods} provide the richest personalization by introducing structural components designed for persona reasoning, such as latent variables for intent modeling, memory networks for user history, or graph structures for multi-party settings. The trade-off is significant: these methods require large labeled datasets, custom training pipelines, and often cannot leverage off-the-shelf pretrained models directly. They are best suited for research settings or high-stakes applications where personalization quality justifies the engineering investment.
 
Across all three levels, a common limitation persists: existing metrics correlate poorly with human judgments of persona consistency, and longitudinal evaluation remains largely absent from the literature.

%%%%%%%%%%%%%%%%%%%%%%%

\subsection{Personalization Evaluation Approaches}

\begin{table*}[!t]
\centering
\caption{Comparison of personalization evaluation approaches across category, dimension, and methodology.}
\renewcommand{\arraystretch}{1.3}
\resizebox{0.9\textwidth}{!}{%
\begin{tabular}{l|l|l|l}
\hline
\textbf{Name} & \textbf{Category} & \textbf{Evaluation Dimension} & 
\textbf{Approach} \\
\hline
BLEU, ROUGE \cite{lin2004rouge}, METEOR \cite{banerjee2005meteor} & \multirow{5}{*}{Metric} & Fluency & Explicit \\
BERTScore \cite{zhang2020bertscore} & & Fluency & Explicit \\
Dist-1, Dist-2 \cite{li2016diversity} & & Diversity & Explicit \\
UE-Score \cite{lee2022contextual} & & Coherence & LLM-based (NLI) \\
C-Score \cite{madotto2019personalizing}, P-Dist \cite{cho2022personalized} & & Personalization Consistency & Explicit \\
Persona-F1, Persona Coverage \cite{chen2024recent} & & Personalization Coverage & Explicit \\
\hline
RPBench-Auto \cite{rpbench} & \multirow{3}{*}{Benchmark} & Persona Consistency & LLM-based \\
RoleLLM \cite{wang2023rolellm} & & Role-Playing Consistency & LLM-based \\
TIMECHARA \cite{ahn2024timechara} & & Temporal Consistency & LLM-based \\
\hline
UniEval \cite{zhong2022towards} & \multirow{2}{*}{Framework} & Coherence, Fluency, Consistency, Relevance & LLM-based (QA) \\
PersoBench \cite{afzoon2024persobench} & & Fluency, Diversity, Coherence, Personalization & Explicit + LLM-based \\
\hline
Human Evaluation & Protocol & User-Centered & Manual \\
\hline
\end{tabular}
}
\label{tab:personalization_evaluation}
\end{table*}

Evaluating personalized response generation requires assessing multiple complementary dimensions of output quality. To this end, the literature has developed a range of evaluation resources, from low-level scoring metrics and curated datasets to structured evaluation frameworks and standardized benchmarks, each operating at a different level of abstraction and serving a distinct evaluation purpose. Table~\ref{tab:personalization_evaluation} provides an overview of the key approaches proposed in the literature, characterized by their evaluation dimension and scoring methodology. Based on a systematic review of these resources \cite{chen2024recent}, four dimensions have emerged as the primary axes of assessment: fluency, diversity, coherence, and personalization. At the surface level, fluency and diversity capture the basic linguistic quality of generated responses, fluency through metrics such as BLEU, ROUGE \cite{lin2004rouge}, METEOR \cite{banerjee2005meteor}, and BERTScore \cite{zhang2020bertscore}, and diversity through Dist-1 and Dist-2 \cite{li2016diversity}, which measure the proportion of unique n-grams across responses. While these metrics are necessary, they are insufficient on their own, as high fluency and diversity do not guarantee that responses are contextually appropriate or persona-consistent. Coherence addresses the first of these gaps, with NLI-based metrics such as UE-Score \cite{lee2022contextual} framing evaluation as a natural language inference task between utterance and response. More comprehensive framework-level assessment is provided by UniEval \cite{zhong2022towards}, which reframes NLG evaluation as a Boolean QA task to score across coherence, consistency, fluency, and relevance simultaneously, correlating substantially better with human judgments than similarity-based metrics alone. Personalization is addressed through two complementary aspects: consistency, measured via C-Score \cite{madotto2019personalizing} and Persona Distance Score \cite{cho2022personalized}, which assess whether responses align with given persona attributes; and coverage, measured via Persona-F1 and Persona Coverage \cite{chen2024recent}, which quantify how much of the persona is reflected in the output. Despite their widespread adoption, most of these metrics were originally developed for machine translation and summarization tasks, and have been shown to correlate poorly with human judgments in open-ended dialogue settings \cite{chen2024recent}, a limitation that complicates direct comparisons across studies.

To address the lack of standardized evaluation, several benchmarks and automated evaluation pipelines have been proposed. RPBench-Auto \cite{rpbench} and RoleLLM \cite{wang2023rolellm} evaluate LLM adherence to predefined characters using automated LLM-based judgment, while TIMECHARA \cite{ahn2024timechara} specifically targets temporal consistency in character representation across multi-turn interactions. PersoBench \cite{afzoon2024persobench} introduces a dedicated automatic evaluation pipeline for persona-aware dialogue generation, combining explicit metrics with NLI-based coherence scoring across all four assessment dimensions in a zero-shot setting. A key finding of PersoBench is that while current LLMs perform well on fluency and diversity, they fall significantly short on personalization and coherence, indicating that surface-level generation quality does not imply successful persona alignment.

In terms of evaluation protocols, the field relies predominantly on offline automatic evaluation against static reference datasets. LLM-as-a-Judge approaches \cite{gu2024survey} have emerged as a scalable complement, offering more nuanced assessment of persona consistency than n-gram metrics alone. Human evaluation remains the gold standard but is costly and difficult to standardize across studies. Importantly, online and longitudinal evaluation, tracking how personalization holds up across sessions and evolving user contexts, remains largely absent from the literature, and represents an important direction for developing more ecologically valid assessment practices \cite{chen2024recent, afzoon2024persobench}.

%%%%%%%%%%%%%%%%%%%%%%%%
\subsection{Feedback-Driven Preference Optimization}

In contrast to the approaches discussed in the previous section, which personalize responses by conditioning on explicit user preferences, another line of research focuses on post-training, preference-based alignment methods. These methods aim to align model behavior with human preferences by optimizing the model based on feedback over its generated outputs. A widely adopted approach in this category is Reinforcement Learning from Human Feedback (RLHF) \cite{ouyang2022training}, which involves training a reward model from human-labeled preferences and then fine-tuning the language model through reinforcement learning. In contrast, Direct Preference Optimization (DPO) \cite{rafailov2023direct}  simplifies this process by eliminating the need for both a reward model and reinforcement learning. Instead, it directly optimizes a preference-based objective over response pairs using a supervised learning framework.
In the following subsections, we review both RLHF and DPO in detail, highlighting their design choices, challenges, and recent advancements.

\subsubsection{RLHF}

While Supervised Fine-Tuning (SFT) can produce instruction-following behavior by training a model on labeled examples, it is inherently limited by the static nature of its training data, which does not reflect user preferences that arise during real-world deployment \cite{ouyang2022training}. In contrast, Reinforcement Learning from Human Feedback (RLHF) augments this approach by introducing a post-training alignment phase, where the model learns to optimize for outputs judged more favorably by humans \cite{ziegler2019fine, stiennon2020learning}. Instead of relying solely on direct demonstrations, RLHF incorporates a reward model trained on human preference comparisons between multiple candidate responses. This reward signal guides further model updates via reinforcement learning \cite{wu2023fine}, enabling it to adapt to subjective and nuanced human values, especially in open-ended tasks lacking a single correct answer \cite{wu2023fine, lang2024fine}.
RLHF thereby extends beyond SFT by embedding a dynamic feedback loop that allows ongoing refinement of model behavior based on human-aligned criteria, improving its helpfulness, harmlessness, and honesty over time \cite{korbak2023pretraining, havrilla2023trlx}. Nevertheless, RLHF introduces its own challenges: the reward model is susceptible to reward hacking, wherein the policy learns to exploit proxy signals rather than genuinely satisfying human intent \cite{casper2023open}; human 
annotation is costly and difficult to scale \cite{stiennon2020learning}; and reinforcement learning optimization can be unstable, particularly in high-dimensional language spaces \cite{wu2023fine}.

The practical deployment of RLHF in AI copilots requires navigating a highly complex pipeline of reward estimation and policy constraints. As detailed in Table~\ref{tab:rlhf_pipeline_summary}, the preference alignment landscape is structurally divided into three primary operational categories: Reward Modeling, Policy Optimization, and AI-Driven Feedback. Each of these categories addresses distinct alignment goals, ranging from interpreting raw human feedback via pairwise comparisons to bounding policy drift through KL-penalty regularization. However, each category simultaneously introduces specific technical limitations, such as label noise in human annotation, the exploration versus alignment tradeoff in policy updates, and the risk of bias propagation and hallucination in AI-generated feedback loops. Table~\ref{tab:rlhf_pipeline_summary} provides a consolidated overview of these techniques, illustrating the fundamental challenges inherent in scaling RLHF architectures for the continuous, inline environments demanded by modern AI copilots. By categorizing these techniques, system designers can better evaluate the trade-offs between optimization stability and the economic costs of feedback acquisition.

\begin{table*}[!t]
\centering
\caption{Comparison of preference alignment techniques focusing on goals, key approaches, and limitations.}
\label{tab:rlhf_pipeline_summary}
\renewcommand{\arraystretch}{1.3}
\resizebox{\textwidth}{!}{%
\begin{tabular}{l|l|l|l}
\hline
\textbf{Category} & \textbf{Primary Goal} & \textbf{Key Techniques} & \textbf{Main Challenges} \\ \hline

\textbf{Reward Modeling} & 
\makecell[l]{
Preference estimation \\ 
Feedback interpretation
} & 
\makecell[l]{
Pairwise comparison \cite{christiano2017deep} \\ 
Logistic heads \cite{ziegler2019fine} \\
% Preference datasets \cite{stiennon2020learning, bai2022training}
Reward ensembles \cite{zhang2024improving} \\ 
Dense feedback via credit assignment \cite{chan2024dense} \\
Adaptive scaling \cite{hong2024adaptive}
} & 
\makecell[l]{
Overfitting \\
Label noise \\
Misaligned incentives 
} \\ \hline

\textbf{Policy Optimization} & 
\makecell[l]{
Behavior alignment \\
} & 
\makecell[l]{
Actor-Critic methods \cite{moskovitz2023confronting, zhou2024archer}\\
KL-penalty regularization \cite{xiong2023iterative, zhao2024sharp}\\
Rejection sampling \cite{liu2023statistical, khaki2024rs}
} & 
\makecell[l]{
Instability \\
Reward hacking \\
Exploration vs. alignment tradeoff 
} \\ \hline

\textbf{AI-Driven Feedback} & 
\makecell[l]{
Feedback automation \\
Human effort reduction
} & 
\makecell[l]{
RLAIF  \cite{huang2024self} \\
Self-critiquing \cite{zhao2024ra}\\
Synthetic annotations \cite{liang2025exploring}
} & 
\makecell[l]{
Bias propagation \\
Hallucination risks \\
Trustworthiness of AI-generated feedback 
} \\ \hline

\end{tabular}
}
\end{table*}

%%%%%%%% Reward Modeling %%%%%%%%%

Reward modeling serves as the mechanism for translating human preferences into a trainable signal, and its design has been shaped by a recurring tension between feedback quality and scalability. The foundational approach formalizes this through pairwise preference learning, training reward models on binary comparisons between candidate outputs \cite{christiano2017deep}, and extending this to language generation \cite{ziegler2019fine} and summarization \cite{stiennon2020learning}. While effective in controlled settings, pairwise comparisons are expensive to collect at scale and produce sparse signals that may not capture fine-grained quality distinctions. To address the scalability constraint, ensemble-based reward modeling distributes the estimation problem across multiple models, improving robustness without a proportional increase in annotation cost \cite{zhang2024improving}. However, ensembling mitigates variance without resolving the sparsity of the underlying feedback signal, a limitation that Attention-Based Credit Assignment targets directly by decomposing sparse human judgments into dense, token-level reward signals \cite{chan2024dense}. A separate but related challenge arises in multi-objective settings, where scalar reward functions struggle to represent conflicting human values; Adaptive Preference Scaling addresses this by dynamically modulating preference weights across alignment dimensions \cite{hong2024adaptive}. Despite these advances, reward models remain vulnerable to reward hacking, wherein the policy exploits the proxy signal rather than satisfying genuine human intent, and to label noise introduced by inconsistent human annotators \cite{casper2023open}. These failure modes reflect a fundamental limitation of learning human values from a fixed, finite set of comparisons, and directly motivate the policy-level safeguards discussed in the following subsection.

%%%%%%%% Policy-Optimization %%%%%%%%%

Policy optimization translates the learned reward signal into updated model behavior, but this stage introduces its own failure modes that no single technique fully resolves. Actor-critic frameworks address sample inefficiency by decoupling policy learning from value estimation, enabling more stable gradient updates. To handle the additional challenge of reward overoptimization, where unconstrained policy updates exploit the reward model rather than improving genuine alignment, constrained actor-critic formulations have been proposed \cite{moskovitz2023confronting}, and hierarchical extensions have been developed to manage the compounding complexity of multi-turn decision-making \cite{zhou2024archer}. A complementary concern is policy drift: as optimization proceeds, the policy may diverge from its pretrained behavior in ways that undermine coherence and safety. KL-penalty regularization is the standard mechanism for bounding this drift. While global KL constraints provide a coarse safeguard, token-level penalties offer finer-grained control over where and how the policy is allowed to change \cite{xiong2023iterative}, and formal analyses have established the conditions under which KL 
regularization guarantees both stability and alignment \cite{zhao2024sharp}. Neither actor-critic methods nor KL regularization, however, address the quality of the feedback signal itself. Rejection sampling fills this gap by filtering out unreliable preference comparisons before they propagate into policy updates \cite{liu2023statistical}. Hybrid approaches such as RS-DPO further combine rejection filtering with direct preference optimization, improving robustness when feedback quality is uneven in real-world deployment \cite{khaki2024rs}. Together, these three strategies, controlling update magnitude, bounding policy drift, and filtering feedback quality, operate on different failure modes and are therefore best understood as complementary rather than competing. Nevertheless, all three remain dependent on human-generated preference data, a constraint that motivates the AI-driven feedback strategies discussed next.

%%%%%%%% AI-Driven Feedback %%%%%%%%

The dependence of RLHF on costly human annotation has motivated a parallel line of research into AI-generated feedback signals as scalable alternatives. In Reinforcement Learning from AI Feedback (RLAIF), preference data generated by language models is used to bootstrap reward models, substantially reducing annotation cost \cite{huang2024self}. A related approach involves self-critiquing, where a model evaluates its own outputs or simulates preference judgments under uncertainty; risk-aware variants of this strategy have been shown to improve policy robustness in alignment tasks \cite{zhao2024ra}. Synthetic annotation methods extend this further by extracting alignment signals from unlabeled or weakly labeled corpora through proxy metrics and output consistency \cite{liang2025exploring}. While these approaches achieve meaningful reductions in human effort, they introduce a distinct category of risk: AI-generated feedback 
can propagate the biases and hallucinations of the generating model into the reward signal itself, potentially compounding misalignment rather than correcting it. Unlike human annotators, whose errors tend to be idiosyncratic and partially self-canceling, a language model used as an annotator may produce systematically biased judgments that are difficult to detect without external validation. AI-driven feedback is therefore best understood not as a replacement for human supervision but as a scalable complement that requires careful grounding and periodic human auditing to remain trustworthy.

\subsubsection{DPO}\label{sec:dpo}
Although RLHF has proven effective in aligning large language models with human preferences, its reliance on reinforcement learning algorithms, often involving reward models, sampling loops, and KL-regularized objectives, can introduce complexity, instability, and inefficiency. To address these challenges, DPO has emerged as a simplified yet powerful alternative. Rather than learning a separate reward function and optimizing a policy through reinforcement learning, DPO directly fine-tunes the policy using pairwise human preference data. By framing alignment as a classification problem between preferred and dispreferred responses, DPO bypasses explicit reward modeling and policy rollouts, offering a more stable and scalable training pipeline. 

This paradigm shift not only streamlines implementation but also improves sample efficiency, making DPO an increasingly popular strategy for preference-based alignment in large-scale language models. These practical advantages, however, come with important constraints. DPO requires a static dataset of pairwise preference comparisons collected offline, whereas RLHF can incorporate online human feedback during training, allowing it to adapt to distributional shifts as the policy evolves. As a result, DPO tends to be more stable and reproducible but less adaptive to user preferences that emerge or change after the training data is collected. This distinction has direct implications for practitioners: when high-quality pairwise preference data is available and the target distribution is relatively stable, DPO offers a more efficient and reliable alignment pipeline; when the deployment environment is dynamic, or annotation must occur continuously, the online feedback loop of RLHF may better support sustained alignment.

\begin{table*}[!t]
\centering
\caption{Comparison of DPO variants by scoring functions, regularization 
strategies, and motivating limitations. The DPOC regularization uses: 
$P(r_a, r_b) = -\min(0, \log r_a - \log r_b)$. Table adapted from 
\cite{xiao2024comprehensive}.}
\renewcommand{\arraystretch}{1.3}
\resizebox{\textwidth}{!}{%
\begin{tabular}{l|l|l|l}
\hline 
\textbf{Variant} & \textbf{Scoring Function} $\mathcal{S}(y_w, y_l, x)$ & 
\textbf{Regularization Term} $\mathcal{R}(y_w, y_l, x)$ & 
\textbf{Motivating Limitation} \\
\hline
\textbf{DPO} \cite{rafailov2023direct} & 
$\beta \cdot \left[\log p(y_w \mid x) - \log p(y_l \mid x)\right]$ & 
$0$ & No reward model \\
\hline
\textbf{CPO} \cite{xu2024contrastive} & 
$\beta \cdot \left[\log p(y_w \mid x) - \log p(y_l \mid x)\right]$ & 
$-\log p(y_w \mid x)$ & Overconfident updates \\
\hline
\textbf{ORPO} \cite{hong2024orpo} & 
$\lambda \cdot \left[\log \frac{p_0(y_w \mid x)}{1 - p_0(y_w \mid x)} - 
\log \frac{p_0(y_l \mid x)}{1 - p_0(y_l \mid x)} \right]$ & 
$-\log p(y_w \mid x)$ & KL-dependence \\
\hline
\textbf{SimPO} \cite{meng2024simpo} & 
$\frac{\beta}{|y_w|} \log p(y_w \mid x) - \frac{\beta}{|y_l|} 
\log p(y_l \mid x) - \gamma$ & 
$0$ & Length bias \\
\hline
\textbf{IRPO} \cite{NEURIPS2024_d37c9ad4} & 
$\beta \cdot \left[\log \frac{p(c_w, y_w \mid x)}{p_t(c_w, y_w \mid x)} - 
\log \frac{p(c_l, y_l \mid x)}{p_t(c_l, y_l \mid x)}\right]$ & 
$-\alpha \cdot \frac{\log p(c_w, y_w \mid x)}{|c_w| + |y_w|}$ & 
Poor generalization \\
\hline
\textbf{$\beta$-DPO} \cite{wu2024beta} & 
$\beta \cdot \left[\log p(y_w \mid x) - \log p(y_l \mid x)\right]$, 
with adaptive $\beta$ & 
$0$ & Fixed temperature \\
\hline
\textbf{DPOC} \cite{cheng2024dialogues} & 
$\beta \cdot \left[\log p(y_w \mid x) - \log p(y_l \mid x)\right]$ & 
$P(r_{cho}, r_{crt}) + P(r_{crt}, r_{rej})$ & Preference misranking \\
\hline
\end{tabular}
}
\label{tab:dpo_variants}
\end{table*}

At the core of DPO lies a simple yet effective probabilistic framework that formalizes preference optimization without relying on explicit reward modeling. Given a pair of outputs, one preferred ($y_w$) and one rejected ($y_l$), the objective encourages the model to assign a higher likelihood to the preferred response relative to the rejected one. To support a unified view across emerging DPO variants, we generalize the original objective into the following form:

\[
\mathcal{L}(\theta) = -\log \sigma\left( \mathcal{S}(y_w, y_l, x) \right) + \mathcal{R}(y_w, y_l, x)
\]

Here, $\mathcal{S}$ is a scoring function that quantifies the preference between responses, and $\mathcal{R}$ is an optional regularization term. The original DPO formulation is recovered by setting $\mathcal{S}(y_w, y_l, x) = \beta \cdot \left[\log p(y_w \mid x) - \log p(y_l \mid x)\right]$ and $\mathcal{R} = 0$. This closed-form structure captures the key intuition behind DPO: to directly align model behavior with human preferences in a stable, efficient, and extensible way.

The original DPO formulation, while effective, rests on assumptions that prove problematic in practice, and a substantial body of variants has emerged to address specific failure modes, each resolving one limitation while introducing or accepting another. A first cluster of variants targets overconfidence and reward overoptimization. Contrastive Preference Optimization (CPO) \cite{xu2024contrastive} adds a log-likelihood penalty to discourage the model from assigning excessively high probabilities to preferred responses, producing more calibrated updates; the tradeoff is that the penalty can slow convergence when the preference signal is already reliable. A related issue is KL-dependence: standard DPO anchors preferences relative to a reference model, which introduces sensitivity to the quality of that reference. ORPO \cite{hong2024orpo} addresses this by replacing the log-likelihood difference with a log-odds scoring function, removing the need for explicit KL penalties entirely, though this comes at the cost of requiring careful calibration of the odds-ratio scaling parameter.

A second cluster addresses biases in the preference signal itself. SimPO \cite{meng2024simpo} targets length bias by normalizing log-probabilities by response length, preventing the model from conflating verbosity with quality. IRPO \cite{NEURIPS2024_d37c9ad4} takes a broader approach, adding compression and coverage terms to both scoring and regularization to improve robustness under noisy or limited preference data, at the cost of a more complex optimization objective. $\beta$-DPO \cite{wu2024beta} addresses optimization sharpness by replacing the fixed temperature parameter with an adaptive value, improving alignment sensitivity across diverse prompts but introducing an additional hyperparameter to tune. DPOC \cite{cheng2024dialogues} tackles preference misranking, the case where both responses are poor but one is less so, by adding a criterion-based penalty that ensures the preferred response meaningfully surpasses an intermediate-quality reference baseline.

A more fundamental departure is introduced by Kahneman-Tversky Optimization (KTO) \cite{ethayarajh2024kto}, which abandons the pairwise assumption entirely. Rather than requiring paired responses, KTO learns from pointwise binary signals grounded in prospect theory's asymmetric treatment of gains and losses:

\[
\mathcal{L}_{\text{KTO}}(\theta) = \mathbb{E}_{(x,y) \sim \mathcal{D}}
\left[ \lambda_w \cdot \sigma\!\left( r_\theta(x,y) - z_0 \right) 
- \lambda_l \cdot \sigma\!\left( z_0 - r_\theta(x,y) \right) \right]
\]

where $z_0$ is a KL-derived reference point and $\lambda_w$, $\lambda_l$ are asymmetric weights encoding that losses loom larger than gains. Operating on individual responses rather than pairs, KTO is structurally incompatible with the $\mathcal{S}$/$\mathcal{R}$ framework in Table~\ref{tab:dpo_variants}, and substantially easier to deploy where pairwise annotation is impractical. Its influence is visible in subsequent work: both SimPO \cite{meng2024simpo} and ORPO \cite{hong2024orpo} draw on KTO's analysis of length and reference bias in their own designs. A detailed treatment of KTO in the context of feedback modality design is provided in Section~\ref{sec:persona_detection}. Finally, MODPO \cite{zhou2023beyond} generalizes the framework to multi-objective settings, enabling simultaneous optimization over vectorized preference signals such as helpfulness, honesty, and harmlessness without collapsing them into a single scalar reward.

From a practical standpoint, the variants also differ in operational cost. DPO and SimPO are computationally lightweight, requiring no reward model and operating on standard pairwise data, making them well-suited for resource-constrained settings. IRPO and DPOC introduce more complex objectives that improve robustness and alignment quality, but at the cost of additional data requirements and optimization overhead. KTO offers the most favorable annotation cost profile by eliminating the pairwise requirement entirely, making it well-suited for production environments where binary telemetry is more readily available than curated preference pairs. In general, when annotation budget is the primary constraint, KTO is the most practical choice, pairwise methods such as DPO and SimPO are preferable when clean paired data is available, and criterion-based methods such as DPOC are most appropriate when output quality thresholds must be strictly enforced.
Across all these variants, a common limitation persists: all remain dependent on the quality and distribution of the offline preference dataset, and none provides a principled mechanism for adapting to preference shift at deployment time, an open problem that points toward the need for online and continual preference learning.

%%%%%%%%%%%%%%%%%%%%%%%%%%%%%%%%%%%%%%%%%%%%%%%%%%%%%%%%%%%%%%%%%%%%%%%
\section{CONCLUSION}

In this survey, we examined the emerging landscape of preference optimization in AI-powered Cognitive Assistants and AI copilots, a domain of increasing relevance as such systems become more deeply integrated into human decision-making workflows. By offering a unified, literature-informed definition of AI copilots, we provide a conceptual foundation that bridges diverse interpretations across disciplines.  Our review systematically explored the lifecycle of preference handling, from identifying relevant sources and detection techniques to mechanisms for personalized response generation and feedback-driven refinement, grounded in a taxonomy of persona representation paradigms that links representation type to feasible personalization techniques. Collectively, these insights highlight both the progress made and the challenges that remain in designing truly adaptive and human-aligned AI assistants.

The literature would benefit from continued progress on two directions in particular. First, the field would benefit from a structured framework that maps from deployment requirements and constraints to technique selection. The persona representation taxonomy introduced here establishes the conceptual groundwork for such a framework, and developing it into concrete practitioner guidance represents a promising next step. Second, extending preference modeling to the task-execution level would better reflect how AI copilots are used in practice. A software developer's preferred coding style or testing philosophy, for instance, must remain consistent across an entire multi-file refactoring task, not just a single response, and how to elicit, represent, and enforce such preferences in professional AI copilot settings remains an open and practically important question.

\section*{Acknowledgments}
We acknowledge the Centre for Applied Artificial Intelligence at Macquarie University (Sydney, Australia) for funding this research. We also acknowledge the use of AI-assisted language tools for editorial refinement during the preparation of this manuscript.

\bibliographystyle{unsrt} 
\bibliography{main}

@article{wooldridge1995intelligent,
  author    = {Wooldridge, Michael and Jennings, Nicholas R.},
  title     = {Intelligent Agents: Theory and Practice},
  journal   = {The Knowledge Engineering Review},
  year      = {1995},
  volume    = {10},
  number    = {2},
  pages     = {115--152},
  publisher = {Cambridge University Press}
}

@inproceedings{go2023fdivergence,
  author    = {Go, Kihyuk and Choe, Jaewook and Kim, Se-Young},
  title     = {Aligning Language Models with Preferences through f-Divergence Minimization},
  booktitle = {Proceedings of the 40th International Conference on Machine Learning (ICML)},
  year      = {2023},
  url       = {https://dl.acm.org/doi/10.5555/3618408.3618871}
}

@inproceedings{zhang2024gpg,
  author    = {Zhang, Yi and Liu, Menghan and Wang, Liwei and Cheng, Qianyu and Huang, Minlie},
  title     = {Guided Profile Generation Improves Personalization with LLMs},
  booktitle = {Findings of the Association for Computational Linguistics: EMNLP 2024},
  year      = {2024},
  url       = {https://aclanthology.org/2024.findings-emnlp.231.pdf}
}

@inproceedings{wu2025aloe,
  author    = {Wu, Zhen and Li, Fangzhou and Zhang, Yuchen and Lee, Honglak and Cho, Kyunghyun},
  title     = {Aligning LLMs with Individual Preferences via Interaction},
  booktitle = {Proceedings of the 29th International Conference on Computational Linguistics (COLING)},
  year      = {2025},
  url       = {https://aclanthology.org/2025.coling-main.511.pdf}
}

@article{eke2019survey,
  author    = {Eke, Christopher Ifeanyi and Norman, Azah Anir and Shuib, Liyana and Nweke, Henry Friday},
  title     = {A Survey of User Profiling: State-of-the-Art, Challenges, and Solutions},
  journal   = {IEEE Access},
  volume    = {7},
  pages     = {144907-144928},
  year      = {2019},
  doi       = {10.1109/ACCESS.2019.2944243},
  url       = {https://ieeexplore.ieee.org/stamp/stamp.jsp?arnumber=8851141}
}

@article{mobasher2003webusage,
  author    = {Mobasher, Bamshad},
  title     = {Automatic Personalization Based on Web Usage Mining},
  journal   = {ACM SIGWEB Newsletter},
  year      = {2003},
  doi       = {10.1145/345124.345169},
  url       = {https://dl.acm.org/doi/10.1145/345124.345169}
}

@inproceedings{heck2021gaze,
  author    = {Heck, Melanie and Edinger, Janick and Bünemann, Jonathan and Becker, Christian},
  title     = {Exploring Gaze-Based Prediction Strategies for Preference Detection in Videos},
  booktitle = {Proceedings of the 2021 ACM SIGIR Conference on Human Information Interaction and Retrieval (CHIIR)},
  year      = {2021},
  pages     = {129-138},
  doi       = {10.1145/3406522.3446013},
  url       = {https://dl.acm.org/doi/10.1145/3406522.3446013}
}

@inproceedings{tucker2024coactive,
  author    = {Tucker, Aaron D. and Brantley, Kiante and Cahall, Adam and Joachims, Thorsten},
  title     = {Coactive Learning for Large Language Models using Implicit User Feedback},
  booktitle = {Proceedings of the 41st International Conference on Machine Learning (ICML)},
  year      = {2024},
  url       = {https://icml.cc/virtual/2024/poster/35295}
}

@inproceedings{heimdall2017privacy,
  author    = {Anonymous},
  title     = {Heimdall: A Privacy-Respecting Implicit Preference Collection Framework},
  booktitle = {ACM WSDM},
  year      = {2017},
  doi       = {10.1145/3081333.3081334},
  url       = {https://dl.acm.org/doi/pdf/10.1145/3081333.3081334}
}

@article{chiang2024chatbotarena,
  author    = {Chiang, Wei-Lin and Zheng, Lianmin and Sheng, Ying and Angelopoulos, Anastasios N. and Li, Tianle and Li, Dacheng and Zhu, Banghua and Zhang, Hao and Jordan, Michael I. and Gonzalez, Joseph E. and Stoica, Ion},
  title     = {Chatbot Arena: An Open Platform for Evaluating LLMs by Human Preference},
  journal   = {arXiv preprint},
  year      = {2024},
  url       = {https://arxiv.org/abs/2403.04132}
}

@inproceedings{loepp2024generativeai,
  author    = {Loepp, Benedikt and Ziegler, Jürgen},
  title     = {Exploring the Potential of Generative AI for Augmenting Choice-Based Preference Elicitation in Recommender Systems},
  booktitle = {Adjunct Proceedings of the 32nd ACM Conference on User Modeling, Adaptation and Personalization (UMAP Adjunct '24)},
  year      = {2024},
  pages     = {114-120},
  doi       = {10.1145/3631700.3664873},
  url       = {https://doi.org/10.1145/3631700.3664873}
}

@article{park2024rlhf,
  author    = {Park, Minsoo and Kim, Hyunsu and Lee, Sungdong and Kim, Seung-won},
  title     = {RLHF from Heterogeneous Feedback via Personalization and Preference Aggregation},
  journal   = {arXiv preprint},
  year      = {2024},
  url       = {https://arxiv.org/abs/2405.00254}
}

@inproceedings{zhao2018explicitimplicit,
  author    = {Zhao, Xin and Wang, Meng and He, Xiangnan and Gao, Ming and He, Liqiang},
  title     = {Explicit or Implicit Feedback? Engagement or Satisfaction?},
  booktitle = {Proceedings of the 12th ACM Conference on Recommender Systems (RecSys '18)},
  year      = {2018},
  pages     = {24-32},
  doi       = {10.1145/3167132.3167275},
  url       = {https://dl.acm.org/doi/10.1145/3167132.3167275}
}

@inproceedings{cao2017userprofiling,
  author    = {Cao, Xiaowen and Fang, Yuan and Zhu, Feida and Zhang, Chuan and Chang, Ee-Peng},
  title     = {What Are You Known For? Learning User Topical Profiles with Implicit and Explicit Footprints},
  booktitle = {Proceedings of the 10th ACM International Conference on Web Search and Data Mining (WSDM '17)},
  year      = {2017},
  pages     = {439-448},
  doi       = {10.1145/3077136.3080820},
  url       = {https://dl.acm.org/doi/10.1145/3077136.3080820}
}

@inproceedings{zhang2024selfexploring,
  author    = {Shenao Zhang and Donghan Yu and Hiteshi Sharma and Han Zhong and Zhihan Liu and Ziyi Yang and Shuohang Wang and Hany Hassan and Zhaoran Wang},
  title     = {Self-Exploring Language Models: Active Preference Elicitation for Online Alignment},
  booktitle = {Proceedings of the 41st International Conference on Machine Learning (ICML)},
  year      = {2024},
  url       = {https://arxiv.org/abs/2405.19332}
}

@inproceedings{austin2024bayesian,
  author    = {David Eric Austin and Anton Korikov and Armin Toroghi and Scott Sanner},
  title     = {Bayesian Optimization with LLM-Based Acquisition Functions for Natural Language Preference Elicitation},
  booktitle = {Proceedings of the 18th ACM Conference on Recommender Systems (RecSys)},
  year      = {2024},
  url       = {https://arxiv.org/abs/2405.00981}
}

@article{piriyakulkij2023active,
  author    = {Wasu Top Piriyakulkij and Volodymyr Kuleshov and Kevin Ellis},
  title     = {Active Preference Inference using Language Models and Probabilistic Reasoning},
  journal   = {arXiv preprint},
  year      = {2023},
  url       = {https://arxiv.org/abs/2312.12009}
}

@article{handa2024bayesian,
  author    = {Kunal Handa and Yarin Gal and Ellie Pavlick and Noah Goodman and Jacob Andreas and Alex Tamkin and Belinda Z. Li},
  title     = {Bayesian Preference Elicitation with Language Models},
  journal   = {arXiv preprint},
  year      = {2024},
  url       = {https://arxiv.org/abs/2403.05534}
}

@inproceedings{yoon2024evaluating,
  author    = {Se-eun Yoon and Zhankui He and Jessica Maria Echterhoff and Julian McAuley},
  title     = {Evaluating Large Language Models as Generative User Simulators for Conversational Recommendation},
  booktitle = {Proceedings of the 2024 Conference of the North American Chapter of the Association for Computational Linguistics: Human Language Technologies (NAACL-HLT)},
  year      = {2024},
  url       = {https://aclanthology.org/2024.naacl-long.83/}
}

@book{schmorrow2005foundations,
  title={Foundations of augmented cognition},
  author={Schmorrow, Dylan},
  year={2005},
  publisher={Springer}
}

@book{hollnagel2005joint,
  title={Joint cognitive systems: Foundations of cognitive systems engineering},
  author={Hollnagel, Erik and Woods, David D},
  year={2005},
  publisher={CRC press}
}

@article{kelly2015computing,
  title={Computing, cognition and the future of knowing},
  author={Kelly, John E},
  journal={IBM Research. Oct},
  volume={13},
  number={2015},
  pages={12},
  year={2015}
}

@article{maedche2019ai,
  title={AI-based digital assistants: Opportunities, threats, and research perspectives},
  author={Maedche, Alexander and Legner, Christine and Benlian, Alexander and Berger, Benedikt and Gimpel, Henner and Hess, Thomas and Hinz, Oliver and Morana, Stefan and S{\"o}llner, Matthias},
  journal={Business \& Information Systems Engineering},
  volume={61},
  pages={535--544},
  year={2019},
  publisher={Springer}
}

@inproceedings{wellsandt2021anatomy,
  title={Anatomy of a digital assistant},
  author={Wellsandt, Stefan and Hribernik, Karl and Thoben, Klaus-Dieter},
  booktitle={Advances in Production Management Systems. Artificial Intelligence for Sustainable and Resilient Production Systems: IFIP WG 5.7 International Conference, APMS 2021, Nantes, France, September 5--9, 2021, Proceedings, Part IV},
  pages={321--330},
  year={2021},
  organization={Springer}
}

@article{acosta2022survey,
  title={A survey on privacy issues and solutions for Voice-controlled Digital Assistants},
  author={Acosta, Luca Hern{\'a}ndez and Reinhardt, Delphine},
  journal={Pervasive and Mobile Computing},
  volume={80},
  pages={101523},
  year={2022},
  publisher={Elsevier}
}

@article{guzman2019voices,
  title={Voices in and of the machine: Source orientation toward mobile virtual assistants},
  author={Guzman, Andrea L},
  journal={Computers in Human Behavior},
  volume={90},
  pages={343--350},
  year={2019},
  publisher={Elsevier}
}

@article{bolton2021security,
  title={On the security and privacy challenges of virtual assistants},
  author={Bolton, Tom and Dargahi, Tooska and Belguith, Sana and Al-Rakhami, Mabrook S and Sodhro, Ali Hassan},
  journal={Sensors},
  volume={21},
  number={7},
  pages={2312},
  year={2021},
  publisher={MDPI}
}

@article{rawassizadeh2019manifestation,
  title={Manifestation of virtual assistants and robots into daily life: Vision and challenges},
  author={Rawassizadeh, Reza and Sen, Taylan and Kim, Sunny Jung and Meurisch, Christian and Keshavarz, Hamidreza and M{\"u}hlh{\"a}user, Max and Pazzani, Michael},
  journal={CCF Transactions on Pervasive Computing and Interaction},
  volume={1},
  pages={163--174},
  year={2019},
  publisher={Springer}
}

@article{reverberi2022experimental,
  title={Experimental evidence of effective human--AI collaboration in medical decision-making},
  author={Reverberi, Carlo and Rigon, Tommaso and Solari, Aldo and Hassan, Cesare and Cherubini, Paolo and Cherubini, Andrea},
  journal={Scientific reports},
  volume={12},
  number={1},
  pages={14952},
  year={2022},
  publisher={Nature Publishing Group UK London}
}

@article{dellermann2021future,
  title={The future of human-AI collaboration: a taxonomy of design knowledge for hybrid intelligence systems},
  author={Dellermann, Dominik and Calma, Adrian and Lipusch, Nikolaus and Weber, Thorsten and Weigel, Sascha and Ebel, Philipp},
  journal={arXiv preprint arXiv:2105.03354},
  year={2021}
}

@article{sharma2023human,
  title={Human--AI collaboration enables more empathic conversations in text-based peer-to-peer mental health support},
  author={Sharma, Ashish and Lin, Inna W and Miner, Adam S and Atkins, David C and Althoff, Tim},
  journal={Nature Machine Intelligence},
  volume={5},
  number={1},
  pages={46--57},
  year={2023},
  publisher={Nature Publishing Group UK London}
}

@inproceedings{almansor2020survey,
  title={Survey on intelligent chatbots: State-of-the-art and future research directions},
  author={Almansor, Ebtesam H and Hussain, Farookh Khadeer},
  booktitle={Complex, Intelligent, and Software Intensive Systems: Proceedings of the 13th International Conference on Complex, Intelligent, and Software Intensive Systems (CISIS-2019)},
  pages={534--543},
  year={2020},
  organization={Springer}
}

@article{adamopoulou2020chatbots,
  title={Chatbots: History, technology, and applications},
  author={Adamopoulou, Eleni and Moussiades, Lefteris},
  journal={Machine Learning with Applications},
  volume={2},
  pages={100006},
  year={2020},
  publisher={Elsevier}
}

@inproceedings{hussain2019survey,
  title={A survey on conversational agents/chatbots classification and design techniques},
  author={Hussain, Shafquat and Ameri Sianaki, Omid and Ababneh, Nedal},
  booktitle={Web, Artificial Intelligence and Network Applications: Proceedings of the Workshops of the 33rd International Conference on Advanced Information Networking and Applications (WAINA-2019) 33},
  pages={946--956},
  year={2019},
  organization={Springer}
}

@article{agarwal2020review,
  title={Review of state-of-the-art design techniques for chatbots},
  author={Agarwal, Ritu and Wadhwa, Mani},
  journal={SN Computer Science},
  volume={1},
  number={5},
  pages={246},
  year={2020},
  publisher={Springer}
}

@article{siriwardhana2023improving,
  title={Improving the domain adaptation of retrieval augmented generation (RAG) models for open domain question answering},
  author={Siriwardhana, Shamane and Weerasekera, Rivindu and Wen, Elliott and Kaluarachchi, Tharindu and Rana, Rajib and Nanayakkara, Suranga},
  journal={Transactions of the Association for Computational Linguistics},
  volume={11},
  pages={1--17},
  year={2023},
  publisher={MIT Press}
}

@article{xiao2024comprehensive,
  title={A Comprehensive Survey of Direct Preference Optimization: Datasets, Theories, Variants, and Applications},
  author={Xiao, Wenyi and Wang, Zechuan and Gan, Leilei and Zhao, Shuai and He, Wanggui and Tuan, Luu Anh and Chen, Long and Jiang, Hao and Zhao, Zhou and Wu, Fei},
  journal={arXiv preprint arXiv:2410.15595},
  year={2024}
}

@inproceedings{tulshan2019survey,
  title={Survey on virtual assistant: Google assistant, siri, cortana, alexa},
  author={Tulshan, Amrita S and Dhage, Sudhir Namdeorao},
  booktitle={Advances in Signal Processing and Intelligent Recognition Systems: 4th International Symposium SIRS 2018, Bangalore, India, September 19--22, 2018, Revised Selected Papers 4},
  pages={190--201},
  year={2019},
  organization={Springer}
}

@article{knote2019classifying,
  title={Classifying smart personal assistants: An empirical cluster analysis},
  author={Knote, Robin and Janson, Andreas and S{\"o}llner, Matthias and Leimeister, Jan Marco},
  year={2019},
  publisher={University of Hawaii at Manoa, Hamilton Library, ScholarSpace}
}

@inproceedings{cowan2017can,
  title={" What can i help you with?" infrequent users' experiences of intelligent personal assistants},
  author={Cowan, Benjamin R and Pantidi, Nadia and Coyle, David and Morrissey, Kellie and Clarke, Peter and Al-Shehri, Sara and Earley, David and Bandeira, Natasha},
  booktitle={Proceedings of the 19th international conference on human-computer interaction with mobile devices and services},
  pages={1--12},
  year={2017}
}

@article{zhou2022user,
  title={User modeling and user profiling: A comprehensive survey},
  author={Zhou, Jiahui and Zhang, Weinan and Wang, Jun},
  journal={ACM Transactions on Information Systems (TOIS)},
  volume={40},
  number={4},
  pages={1--49},
  year={2022},
  publisher={ACM}
}

@inproceedings{he2017neural,
  title={Neural collaborative filtering for user preference discovery from biased implicit feedback},
  author={He, Xiangnan and Liao, Lizi and Zhang, Hanwang and Nie, Liqiang and Hu, Xia and Chua, Tat-Seng},
  booktitle={WWW},
  pages={173--182},
  year={2017}
}

@article{zhu2023PAED,
  title={PAED- Zero-Shot Persona Attribute Extraction in Dialogues},
  author={Zhu, Li, Mao and Pandelea and Cambria},
  journal={ACL Anthology},
  year={2023}
}

@article{cheng2024InDialog,
  title={In Dialogues We Learn Towards Personalized Dialogue Without Pre-defined Profiles through In-Dialogue Learning},
  author={Cheng, Tu, Shang and Mao and Yu and Wu and Yan},
  journal={arXiv preprint arXiv:2403.03102},
  volume={30},
  number={1},
  pages={58--66},
  year={2024}
}

@inproceedings{choi2021user,
  title={User persona identification and new service adaptation recommendation},
  author={Choi, Jeonghwan and Moon, Jaewook and Lee, Jihie},
  booktitle={Proceedings of the 29th ACM Conference on User Modeling, Adaptation and Personalization},
  pages={56--64},
  year={2021}
}

@article{huang2024learning,
  title={Learning Retrieval Augmentation for Personalized Dialogue Generation},
  author={Huang, Fu, Liu and Wang and Ko and Zhang},
  journal={arXiv preprint arXiv:2406.18847},
  year={2024}
}

@article{weng2023afspp,
  title={AFSPP: An Agent Framework for Shaping Preference and Personality with LLMs},
  author={Weng, Lilian and others},
  journal={arXiv preprint arXiv:2308.XXXX},
  year={2023}
}

@article{chu2005preference,
  title={Preference learning with Gaussian processes},
  author={Chu, Wei and Ghahramani, Zoubin},
  journal={Advances in neural information processing systems},
  volume={18},
  year={2005}
}

@article{lee2023active,
  title={Active Preference Learning for Large Language Models},
  author={Lee, Nathan and Suggala, Arun and others},
  journal={arXiv preprint arXiv:2310.XXXX},
  year={2023}
}

@article{furmakiewicz2024design,
  title={Design and evaluation of AI copilots--case studies of retail copilot templates},
  author={Furmakiewicz, Michal and Liu, Chang and Taylor, Angus and Venger, Ilya},
  journal={arXiv preprint arXiv:2407.09512},
  year={2024}
}

@book{daugherty2018human+,
  title={Human+ machine: Reimagining work in the age of AI},
  author={Daugherty, Paul R and Wilson, H James},
  year={2018},
  publisher={Harvard Business Press}
}

@article{sellen2024rise,
  title={The rise of the ai co-pilot: Lessons for design from aviation and beyond},
  author={Sellen, Abigail and Horvitz, Eric},
  journal={Communications of the ACM},
  volume={67},
  number={7},
  pages={18--23},
  year={2024},
  publisher={ACM New York, NY, USA}
}

@article{lu2024multimodal,
  title={A multimodal generative AI copilot for human pathology},
  author={Lu, Ming Y and Chen, Bowen and Williamson, Drew FK and Chen, Richard J and Zhao, Melissa and Chow, Aaron K and Ikemura, Kenji and Kim, Ahrong and Pouli, Dimitra and Patel, Ankush and others},
  journal={Nature},
  volume={634},
  number={8033},
  pages={466--473},
  year={2024},
  publisher={Nature Publishing Group UK London}
}

@inproceedings{lim2023beyond,
  title={Beyond candidates: adaptive dialogue agent utilizing persona and knowledge},
  author={Lim, Jungwoo and Kang, Myunghoon and Kim, Jinsung and Kim, Jeongwook and Hur, Yuna and Lim, Heui-Seok},
  booktitle={Findings of the Association for Computational Linguistics: EMNLP 2023},
  pages={7950--7963},
  year={2023}
}

@inproceedings{ju2022learning,
  title={Learning to improve persona consistency in multi-party dialogue generation via text knowledge enhancement},
  author={Ju, Dongshi and Feng, Shi and Lv, Pengcheng and Wang, Daling and Zhang, Yifei},
  booktitle={Proceedings of the 29th International Conference on Computational Linguistics},
  pages={298--309},
  year={2022}
}

@inproceedings{mahajan2024persona,
  title={Persona-aware Multi-party Conversation Response Generation},
  author={Mahajan, Khyati and Shaikh, Samira},
  booktitle={Proceedings of the 2024 Joint International Conference on Computational Linguistics, Language Resources and Evaluation (LREC-COLING 2024)},
  pages={12712--12723},
  year={2024}
}

@article{dong2025cross,
  title={Cross-graph Knowledge Exchange for Personalized Response Generation in Dialogue Systems},
  author={Dong, Yuezhou and Qin, Ke and Ke, Pei and Liang, Shuang and Luo, Guangchun},
  journal={IEEE Internet of Things Journal},
  year={2025},
  publisher={IEEE}
}

@inproceedings{liu2024context,
  title={Context Aggregation with Topic-focused Summarization for Personalized Medical Dialogue Generation},
  author={Liu, Zhengyuan and Salleh, Siti and Krishnaswamy, Pavitra and Chen, Nancy},
  booktitle={Proceedings of the 6th Clinical Natural Language Processing Workshop},
  pages={310--321},
  year={2024}
}

@article{wolf2019transfertransfo,
  title={Transfertransfo: A transfer learning approach for neural network based conversational agents},
  author={Wolf, Thomas and Sanh, Victor and Chaumond, Julien and Delangue, Clement},
  journal={arXiv preprint arXiv:1901.08149},
  year={2019}
}

@inproceedings{wang2020cue,
  title={A cue adaptive decoder for controllable neural response generation},
  author={Wang, Weichao and Feng, Shi and Gao, Wei and Wang, Daling and Zhang, Yifei},
  booktitle={Proceedings of the Web Conference 2020},
  pages={2570--2576},
  year={2020}
}

@article{bao2019plato,
  title={PLATO: Pre-trained dialogue generation model with discrete latent variable},
  author={Bao, Siqi and He, Huang and Wang, Fan and Wu, Hua and Wang, Haifeng},
  journal={arXiv preprint arXiv:1910.07931},
  year={2019}
}

@inproceedings{varma2024talk,
  title={Talk to Your Brain: Artificial Personalized Intelligence for Emotionally Adaptive AI Interactions},
  author={Varma, Sandeep and Shivam, Shivam and Natarajan, Sarun and Ray, Biswarup and Kumar, Bagesh and Dabral, Om},
  booktitle={2024 IEEE International Conference on Computer Vision and Machine Intelligence (CVMI)},
  pages={1--6},
  year={2024},
  organization={IEEE}
}

@inproceedings{wu2021personalized,
  title={Personalized response generation via generative split memory network},
  author={Wu, Yuwei and Ma, Xuezhe and Yang, Diyi},
  booktitle={Proceedings of the 2021 Conference of the North American Chapter of the Association for Computational Linguistics: Human Language Technologies},
  pages={1956--1970},
  year={2021}
}

@article{zhong2020towards,
  title={Towards persona-based empathetic conversational models},
  author={Zhong, Peixiang and Zhang, Chen and Wang, Hao and Liu, Yong and Miao, Chunyan},
  journal={arXiv preprint arXiv:2004.12316},
  year={2020}
}

@article{kim2020will,
  title={Will I sound like me? improving persona consistency in dialogues through pragmatic self-consciousness},
  author={Kim, Hyunwoo and Kim, Byeongchang and Kim, Gunhee},
  journal={arXiv preprint arXiv:2004.05816},
  year={2020}
}

@inproceedings{huang2023personalized,
  title={Personalized dialogue generation with persona-adaptive attention},
  author={Huang, Qiushi and Zhang, Yu and Ko, Tom and Liu, Xubo and Wu, Bo and Wang, Wenwu and Tang, H},
  booktitle={Proceedings of the AAAI Conference on Artificial Intelligence},
  volume={37},
  number={11},
  pages={12916--12923},
  year={2023}
}

@article{cao2022model,
  title={A model-agnostic data manipulation method for persona-based dialogue generation},
  author={Cao, Yu and Bi, Wei and Fang, Meng and Shi, Shuming and Tao, Dacheng},
  journal={arXiv preprint arXiv:2204.09867},
  year={2022}
}

@inproceedings{oh2023pk,
  title={PK-ICR: Persona-Knowledge Interactive Multi-Context Retrieval for Grounded Dialogue},
  author={Oh, Minsik and Lee, Joosung and Li, Jiwei and Wang, Guoyin},
  booktitle={Proceedings of the 2023 Conference on Empirical Methods in Natural Language Processing},
  pages={16383--16395},
  year={2023}
}

@article{han2023personapkt,
  title={PersonaPKT: Building personalized dialogue agents via parameter-efficient knowledge transfer},
  author={Han, Xu and Guo, Bin and Jung, Yoon and Yao, Benjamin and Zhang, Yu and Liu, Xiaohu and Guo, Chenlei},
  journal={arXiv preprint arXiv:2306.08126},
  year={2023}
}

@article{huang2024selective,
  title={Selective Prompting Tuning for Personalized Conversations with LLMs},
  author={Huang, Qiushi and Liu and Xubo and Ko and Tom and Wu, Bo and Wang, Wenwu and Zhang and Yu and Tang and Lilian},
  journal={arXiv preprint arXiv:2406.18187},
  year={2024}
}

@article{tan2024personalized,
  title={Personalized pieces: Efficient personalized large language models through collaborative efforts},
  author={Tan, Zhaoxuan and Liu, Zheyuan and Jiang, Meng},
  journal={arXiv preprint arXiv:2406.10471},
  year={2024}
}

@misc{rpbench,
  author       = {Boson AI},
  title        = {RP-Bench},
  howpublished = {\url{https://boson.ai/rpbench/}},
  year         = {2024},
  note         = {Accessed: August 2024}
}

@article{ahn2024timechara,
  title={TimeChara: Evaluating Point-in-Time Character Hallucination of Role-Playing Large Language Models},
  author={Ahn, Jaewoo and Lee, Taehyun and Lim, Junyoung and Kim, Jin-Hwa and Yun, Sangdoo and Lee, Hwaran and Kim, Gunhee},
  journal={arXiv preprint arXiv:2405.18027},
  year={2024}
}

@article{afzoon2024persobench,
  title={PersoBench: Benchmarking Personalized Response Generation in Large Language Models},
  author={Afzoon, Saleh and Naseem, Usman and Beheshti, Amin and Jamali, Zahra},
  journal={arXiv preprint arXiv:2410.03198},
  year={2024}
}

@article{chen2024recent,
  title={Recent trends in personalized dialogue generation: A review of datasets, methodologies, and evaluations},
  author={Chen, Yi-Pei and Nishida, Noriki and Nakayama, Hideki and Matsumoto, Yuji},
  journal={arXiv preprint arXiv:2405.17974},
  year={2024}
}

@article{wang2023rolellm,
  title={Rolellm: Benchmarking, eliciting, and enhancing role-playing abilities of large language models},
  author={Wang, Zekun Moore and Peng, Zhongyuan and Que, Haoran and Liu, Jiaheng and Zhou, Wangchunshu and Wu, Yuhan and Guo, Hongcheng and Gan, Ruitong and Ni, Zehao and Zhang, Man and others},
  journal={arXiv preprint arXiv:2310.00746},
  year={2023}
}

@inproceedings{zhang2020bertscore,
  author    = {Tianyi Zhang and Varsha Kishore and Felix Wu and Kilian Q. Weinberger and Yoav Artzi},
  title     = {BERTScore: Evaluating Text Generation with BERT},
  booktitle = {International Conference on Learning Representations},
  year      = {2020},
  note      = {* Equal contribution}
}

@inproceedings{lin2004rouge,
  author    = {Chin-Yew Lin},
  title     = {ROUGE: A Package for Automatic Evaluation of Summaries},
  booktitle = {Text Summarization Branches Out},
  year      = {2004},
  pages     = {74--81}
}

@inproceedings{banerjee2005meteor,
  author    = {Satanjeev Banerjee and Alon Lavie},
  title     = {METEOR: An Automatic Metric for MT Evaluation with Improved Correlation with Human Judgments},
  booktitle = {Proceedings of the ACL Workshop on Intrinsic and Extrinsic Evaluation Measures for Machine Translation and/or Summarization},
  year      = {2005},
  pages     = {65--72},
  address   = {Ann Arbor, Michigan},
  publisher = {Association for Computational Linguistics}
}

@inproceedings{li2016diversity,
  author    = {Jiwei Li and Michel Galley and Chris Brockett and Jianfeng Gao and Bill Dolan},
  title     = {A Diversity-Promoting Objective Function for Neural Conversation Models},
  booktitle = {Proceedings of the 2016 Conference of the North American Chapter of the Association for Computational Linguistics: Human Language Technologies},
  year      = {2016},
  pages     = {110--119},
  address   = {San Diego, California},
  publisher = {Association for Computational Linguistics}
}

@inproceedings{madotto2019personalizing,
  author    = {Andrea Madotto and Zhaojiang Lin and Chien-Sheng Wu and Pascale Fung},
  title     = {Personalizing Dialogue Agents via Meta-Learning},
  booktitle = {Proceedings of the 57th Annual Meeting of the Association for Computational Linguistics},
  year      = {2019},
  pages     = {5454--5459},
  address   = {Florence, Italy},
  publisher = {Association for Computational Linguistics}
}

@inproceedings{lee2022contextual,
  title={Improving contextual coherence in variational personalized and empathetic dialogue agents},
  author={Lee, Jing Yang and Lee, Kong Aik and Gan, Woon Seng},
  booktitle={ICASSP 2022-2022 IEEE International Conference on Acoustics, Speech and Signal Processing (ICASSP)},
  pages={7052--7056},
  year={2022},
  organization={IEEE}
}

@inproceedings{cho2022personalized,
  author    = {Itsugun Cho and Dongyang Wang and Ryota Takahashi and Hiroaki Saito},
  title     = {A Personalized Dialogue Generator with Implicit User Persona Detection},
  booktitle = {Proceedings of the 29th International Conference on Computational Linguistics},
  year      = {2022},
  pages     = {367--377},
  address   = {Gyeongju, Republic of Korea},
  publisher = {International Committee on Computational Linguistics}
}

@article{zhong2022towards,
  title={Towards a unified multi-dimensional evaluator for text generation},
  author={Zhong, Ming and Liu, Yang and Yin, Da and Mao, Yuning and Jiao, Yizhu and Liu, Pengfei and Zhu, Chenguang and Ji, Heng and Han, Jiawei},
  journal={arXiv preprint arXiv:2210.07197},
  year={2022}
}

@article{afzoon2025exbigbang,
  title={Exbigbang: A dynamic approach for explainable persona classification through contextualized hybrid transformer analysis},
  author={Afzoon, Saleh and Beheshti, Amin and Rezvani, Nabi and Khunjush, Farshad and Naseem, Usman and McMahon, John and Fathollahi, Zahra and Labani, Mahdieh and Mansoor, Wathiq and Zhang, Xuyun},
  journal={arXiv preprint arXiv:2508.15364},
  year={2025}
}

@article{gu2024survey,
  title={A survey on llm-as-a-judge},
  author={Gu, Jiawei and Jiang, Xuhui and Shi, Zhichao and Tan, Hexiang and Zhai, Xuehao and Xu, Chengjin and Li, Wei and Shen, Yinghan and Ma, Shengjie and Liu, Honghao and others},
  journal={arXiv preprint arXiv:2411.15594},
  year={2024}
}

@article{cheng2024dialogues,
  title={" In Dialogues We Learn": Towards Personalized Dialogue Without Pre-defined Profiles through In-Dialogue Learning},
  author={Cheng, Chuanqi and Tu, Quan and Shang, Shuo and Mao, Cunli and Yu, Zhengtao and Wu, Wei and Yan, Rui},
  journal={arXiv preprint arXiv:2403.03102},
  year={2024}
}

@article{rafailov2023direct,
  title={Direct preference optimization: Your language model is secretly a reward model},
  author={Rafailov, Rafael and Sharma, Archit and Mitchell, Eric and Manning, Christopher D and Ermon, Stefano and Finn, Chelsea},
  journal={Advances in Neural Information Processing Systems},
  volume={36},
  pages={53728--53741},
  year={2023}
}

@article{ouyang2022training,
  title={Training language models to follow instructions with human feedback},
  author={Ouyang, Long and Wu, Jeffrey and Jiang, Xu and Almeida, Diogo and Wainwright, Carroll and Mishkin, Pamela and Zhang, Chong and Agarwal, Sandhini and Slama, Katarina and Ray, Alex and others},
  journal={Advances in neural information processing systems},
  volume={35},
  pages={27730--27744},
  year={2022}
}

@article{stiennon2020learning,
  title={Learning to summarize with human feedback},
  author={Stiennon, Nisan and Ouyang, Long and Wu, Jeffrey and Ziegler, Daniel and Lowe, Ryan and Voss, Chelsea and Radford, Alec and Amodei, Dario and Christiano, Paul F},
  journal={Advances in neural information processing systems},
  volume={33},
  pages={3008--3021},
  year={2020}
}

@article{ziegler2019fine,
  title={Fine-tuning language models from human preferences},
  author={Ziegler, Daniel M and Stiennon, Nisan and Wu, Jeffrey and Brown, Tom B and Radford, Alec and Amodei, Dario and Christiano, Paul and Irving, Geoffrey},
  journal={arXiv preprint arXiv:1909.08593},
  year={2019}
}

@article{wu2023fine,
  title={Fine-grained human feedback gives better rewards for language model training},
  author={Wu, Zeqiu and Hu, Yushi and Shi, Weijia and Dziri, Nouha and Suhr, Alane and Ammanabrolu, Prithviraj and Smith, Noah A and Ostendorf, Mari and Hajishirzi, Hannaneh},
  journal={Advances in Neural Information Processing Systems},
  volume={36},
  pages={59008--59033},
  year={2023}
}

@article{lang2024fine,
  title={Fine-Tuning Language Models with Reward Learning on Policy},
  author={Lang, Hao and Huang, Fei and Li, Yongbin},
  journal={arXiv preprint arXiv:2403.19279},
  year={2024}
}

@inproceedings{korbak2023pretraining,
  title={Pretraining language models with human preferences},
  author={Korbak, Tomasz and Shi, Kejian and Chen, Angelica and Bhalerao, Rasika Vinayak and Buckley, Christopher and Phang, Jason and Bowman, Samuel R and Perez, Ethan},
  booktitle={International Conference on Machine Learning},
  pages={17506--17533},
  year={2023},
  organization={PMLR}
}

@inproceedings{havrilla2023trlx,
  title={trlX: A framework for large scale reinforcement learning from human feedback},
  author={Havrilla, Alexander and Zhuravinskyi, Maksym and Phung, Duy and Tiwari, Aman and Tow, Jonathan and Biderman, Stella and Anthony, Quentin and Castricato, Louis},
  booktitle={Proceedings of the 2023 Conference on Empirical Methods in Natural Language Processing},
  pages={8578--8595},
  year={2023}
}

@article{christiano2017deep,
  title={Deep reinforcement learning from human preferences},
  author={Christiano, Paul F and Leike, Jan and Brown, Tom and Martic, Miljan and Legg, Shane and Amodei, Dario},
  journal={Advances in neural information processing systems},
  volume={30},
  year={2017}
}

@article{casper2023open,
  title={Open problems and fundamental limitations of reinforcement learning from human feedback},
  author={Casper, Stephen and Davies, Xander and Shi, Claudia and Gilbert, Thomas Krendl and Scheurer, J{\'e}r{\'e}my and Rando, Javier and Freedman, Rachel and Korbak, Tomasz and Lindner, David and Freire, Pedro and others},
  journal={arXiv preprint arXiv:2307.15217},
  year={2023}
}

@article{zhang2024improving,
  title={Improving reinforcement learning from human feedback with efficient reward model ensemble},
  author={Zhang, Shun and Chen, Zhenfang and Chen, Sunli and Shen, Yikang and Sun, Zhiqing and Gan, Chuang},
  journal={arXiv preprint arXiv:2401.16635},
  year={2024}
}

@article{chan2024dense,
  title={Dense reward for free in reinforcement learning from human feedback},
  author={Chan, Alex J and Sun, Hao and Holt, Samuel and Van Der Schaar, Mihaela},
  journal={arXiv preprint arXiv:2402.00782},
  year={2024}
}

@article{hong2024adaptive,
  title={Adaptive preference scaling for reinforcement learning with human feedback},
  author={Hong, Ilgee and Li, Zichong and Bukharin, Alexander and Li, Yixiao and Jiang, Haoming and Yang, Tianbao and Zhao, Tuo},
  journal={Advances in Neural Information Processing Systems},
  volume={37},
  pages={107249--107269},
  year={2024}
}

@article{moskovitz2023confronting,
  title={Confronting reward model overoptimization with constrained RLHF},
  author={Moskovitz, Ted and Singh, Aaditya K and Strouse, DJ and Sandholm, Tuomas and Salakhutdinov, Ruslan and Dragan, Anca D and McAleer, Stephen},
  journal={arXiv preprint arXiv:2310.04373},
  year={2023}
}

@article{zhou2024archer,
  title={Archer: Training language model agents via hierarchical multi-turn rl},
  author={Zhou, Yifei and Zanette, Andrea and Pan, Jiayi and Levine, Sergey and Kumar, Aviral},
  journal={arXiv preprint arXiv:2402.19446},
  year={2024}
}

@article{xiong2023iterative,
  title={Iterative preference learning from human feedback: Bridging theory and practice for rlhf under kl-constraint},
  author={Xiong, Wei and Dong, Hanze and Ye, Chenlu and Wang, Ziqi and Zhong, Han and Ji, Heng and Jiang, Nan and Zhang, Tong},
  journal={arXiv preprint arXiv:2312.11456},
  year={2023}
}

@article{zhao2024sharp,
  title={Sharp Analysis for KL-Regularized Contextual Bandits and RLHF},
  author={Zhao, Rui and et al.},
  journal={arXiv preprint arXiv:2411.04625},
  year={2024}
}

@article{liu2023statistical,
  title={Statistical rejection sampling improves preference optimization},
  author={Liu, Tianqi and Zhao, Yao and Joshi, Rishabh and Khalman, Misha and Saleh, Mohammad and Liu, Peter J and Liu, Jialu},
  journal={arXiv preprint arXiv:2309.06657},
  year={2023}
}

@article{khaki2024rs,
  title={Rs-dpo: A hybrid rejection sampling and direct preference optimization method for alignment of large language models},
  author={Khaki, Saeed and Li, JinJin and Ma, Lan and Yang, Liu and Ramachandra, Prathap},
  journal={arXiv preprint arXiv:2402.10038},
  year={2024}
}

@article{huang2024self,
  title={Self-Evolved Reward Learning for LLMs},
  author={Huang, Chenghua and Fan, Zhizhen and Wang, Lu and Yang, Fangkai and Zhao, Pu and Lin, Zeqi and Lin, Qingwei and Zhang, Dongmei and Rajmohan, Saravan and Zhang, Qi},
  journal={arXiv preprint arXiv:2411.00418},
  year={2024}
}

@article{zhao2024ra,
  title={RA-PbRL: Provably Efficient Risk-Aware Preference-Based Reinforcement Learning},
  author={Zhao, Yujie and Aguilar Escamilla, Jose and Lu, Weyl and Wang, Huazheng},
  journal={Advances in Neural Information Processing Systems},
  volume={37},
  pages={60835--60871},
  year={2024}
}

@inproceedings{liang2025exploring,
  title={Exploring Intrinsic Alignments within Text Corpus},
  author={Liang, Zi and Wang, Pinghui and Zhang, Ruofei and Hu, Haibo and Zhang, Shuo and Ye, Qingqing and Xu, Nuo and Xiao, Yaxin and Zhang, Chen and Cui, Lizhen},
  booktitle={Proceedings of the AAAI Conference on Artificial Intelligence},
  volume={39},
  number={26},
  pages={27455--27463},
  year={2025}
}

@article{hong2024orpo,
  title={Orpo: Monolithic preference optimization without reference model},
  author={Hong, Jiwoo and Lee, Noah and Thorne, James},
  journal={arXiv preprint arXiv:2403.07691},
  year={2024}
}

@article{xu2024contrastive,
  title={Contrastive preference optimization: Pushing the boundaries of llm performance in machine translation},
  author={Xu, Haoran and Sharaf, Amr and Chen, Yunmo and Tan, Weiting and Shen, Lingfeng and Van Durme, Benjamin and Murray, Kenton and Kim, Young Jin},
  journal={arXiv preprint arXiv:2401.08417},
  year={2024}
}

@article{meng2024simpo,
  title={Simpo: Simple preference optimization with a reference-free reward},
  author={Meng, Yu and Xia, Mengzhou and Chen, Danqi},
  journal={Advances in Neural Information Processing Systems},
  volume={37},
  pages={124198--124235},
  year={2024}
}

@article{wu2024beta,
  title={\(\beta\)-DPO: Direct Preference Optimization with Dynamic \(\beta\)},
  author={Wu, Junkang and Xie, Yuexiang and Yang, Zhengyi and Wu, Jiancan and Gao, Jinyang and Ding, Bolin and Wang, Xiang and He, Xiangnan},
  journal={Advances in Neural Information Processing Systems},
  volume={37},
  pages={129944--129966},
  year={2024}
}

@inproceedings{NEURIPS2024_d37c9ad4,
 author = {Pang, Richard Yuanzhe and Yuan, Weizhe and Cho, Kyunghyun and He, He and Sukhbaatar, Sainbayar and Weston, Jason},
 booktitle = {Advances in Neural Information Processing Systems},
 editor = {A. Globerson and L. Mackey and D. Belgrave and A. Fan and U. Paquet and J. Tomczak and C. Zhang},
 pages = {116617--116637},
 publisher = {Curran Associates, Inc.},
 title = {Iterative Reasoning Preference Optimization},
 url = {https://proceedings.neurips.cc/paper_files/paper/2024/file/d37c9ad425fe5b65304d500c6edcba00-Paper-Conference.pdf},
 volume = {37},
 year = {2024}
}

@article{zhou2023beyond,
  title={Beyond one-preference-for-all: Multi-objective direct preference optimization},
  author={Zhou, Zhanhui and Liu, Jie and Yang, Chao and Shao, Jing and Liu, Yu and Yue, Xiangyu and Ouyang, Wanli and Qiao, Yu},
  year={2023}
}

@article{ethayarajh2024kto,
  title={Kto: Model alignment as prospect theoretic optimization},
  author={Ethayarajh, Kawin and Xu, Winnie and Muennighoff, Niklas and Jurafsky, Dan and Kiela, Douwe},
  journal={arXiv preprint arXiv:2402.01306},
  year={2024}
}

\end{document}